\documentclass[11pt]{article}

\usepackage[preprint]{acl}

\usepackage{times}
\usepackage{latexsym}

\usepackage[T1]{fontenc}

\usepackage[utf8]{inputenc}

\usepackage{microtype}

\usepackage{inconsolata}

\usepackage{graphicx}

\usepackage{amsmath}      %
\usepackage{amssymb}
\usepackage{booktabs}     %
\usepackage{enumitem}     %
\usepackage{colortbl}
\usepackage{multirow}
\usepackage{tabularx}
\usepackage{placeins}
\newcolumntype{R}{>{\raggedleft\arraybackslash}X}
\usepackage[most]{tcolorbox}
\tcbset{modprompt/.style={
  breakable, enhanced,  %
  colback=gray!5, colframe=black!80, boxrule=0.6pt, arc=1pt,
  left=5pt, right=5pt, top=4pt, bottom=4pt,
  fonttitle=\bfseries, coltitle=black, colbacktitle=gray!20,
}}
\usepackage{bm}
\usepackage{tikz}         %
\usetikzlibrary{arrows.meta,positioning,fit,calc}

\newcommand{\embmodel}{\texttt{text-\allowbreak embedding-\allowbreak 3-\allowbreak large}}
\newcommand{\judgemodel}{\texttt{gpt-4o-\allowbreak mini}}
\newcommand{\fnvone}{\texttt{check\_\allowbreak idea\_\allowbreak novelty}}
\newcommand{\fncoi}{\texttt{check\_novel}}
\newcommand{\fnsi}{\texttt{retrieve\_\allowbreak novelty\_\allowbreak score}}

\usepackage{xcolor}
\definecolor{srcblue}{HTML}{205898}
\definecolor{srcbluebg}{HTML}{D8E0E8}
\definecolor{okgreen}{HTML}{387820}
\definecolor{okgreenbg}{HTML}{D8F0D0}
\definecolor{badred}{HTML}{C00000}
\definecolor{badredbg}{HTML}{F0D8D8}
\definecolor{tagbg}{HTML}{F0F0F0}
\newcommand{\dimtag}[1]{\colorbox{tagbg}{\scriptsize\textsf{#1}}}
\newcommand{\verdictbox}[3]{\colorbox{#1}{\textcolor{#2}{\textbf{#3}}}}
\newcommand{\src}[1]{\textcolor{srcblue}{\textbf{#1}}}
\newif\ifdraftnotes
\draftnotesfalse

\definecolor{tentcol}{RGB}{170,85,0}    %
\definecolor{todocol}{RGB}{190,30,45}   %

\newif\ifmarkchanges
\markchangesfalse
\definecolor{claudecol}{RGB}{0,90,200}    %
\definecolor{chengcol}{RGB}{140,40,170}   %
\definecolor{collabcol}{RGB}{0,130,70}    %
\ifmarkchanges
  \newcommand{\mine}[1]{{\color{claudecol}#1}}

\else
  \newcommand{\mine}[1]{#1}

\fi

\ifdraftnotes
  
  \newcommand{\td}[1]{{\color{todocol}\textsf{\small[\textbf{TODO} #1]}}}
  
\else
  
  \newcommand{\td}[1]{}
  
\fi

\title{Do AI Scientists Know What's New?}

\author{
 \textbf{Miri Liu}
 \quad
 \textbf{Xiayu Zhao}
 \quad
 \textbf{Encheng Xie}
 \quad
 \textbf{ChengXiang Zhai}
\\
\\
 University of Illinois, Urbana-Champaign
\\
 \texttt{\{miri3, xiayuz3, encheng2, czhai\}@illinois.edu}
}

\begin{document}
\maketitle
\begin{abstract}
A major challenge facing AI scientific discovery pipelines is how to determine when a generated idea is meaningfully novel. Ideally, novelty checks in these pipelines must be able to reject non-novel ideas while \textit{not} rejecting too many novel ideas, and they should also be able to distinguish which parts of a proposal are non-novel, as opposed to opaquely discarding entire proposals. We design an analytical framework for evaluation of novelty checks with temporal ground truth at both the proposal- and claim-level, using four datasets of 1{,}060 proposals built from published papers, reviewer disputes, and generated ideas matching ones the field went on to publish. Auditing existing novelty checks with this framework, we find that checks with high rejection rates on non-novel work also tend to have high false alarms on ideas that are later realized by human researchers, and observe that checks built to evaluate proposals as a whole can point to claim-level prior art only by rejecting the whole proposal. We therefore introduce \textsc{ClaimNov}, a fine-grained, interpretable claim-level novelty module. \textsc{ClaimNov} catches 97\% of disguised papers at a 1.5\% false-alarm rate, and its claim-level novelty profile cuts the LLM judge's false alarms from 11.8\% to 1.5\%. Because its verdict rests on claim-level evidence, \textsc{ClaimNov} also keeps proposals that build on established contributions, rejecting only 3\% of ICLR-accepted proposals with a reviewer-disputed contribution and keeping 91 of 100 novel proposals with one spliced prior contribution, against 76 for the strongest existing check. Code and data are available at \url{https://github.com/emmatliu/claimnov}.
\end{abstract}

\section{Introduction}

\begin{figure}[!t]
\centering
\includegraphics[width=\columnwidth]{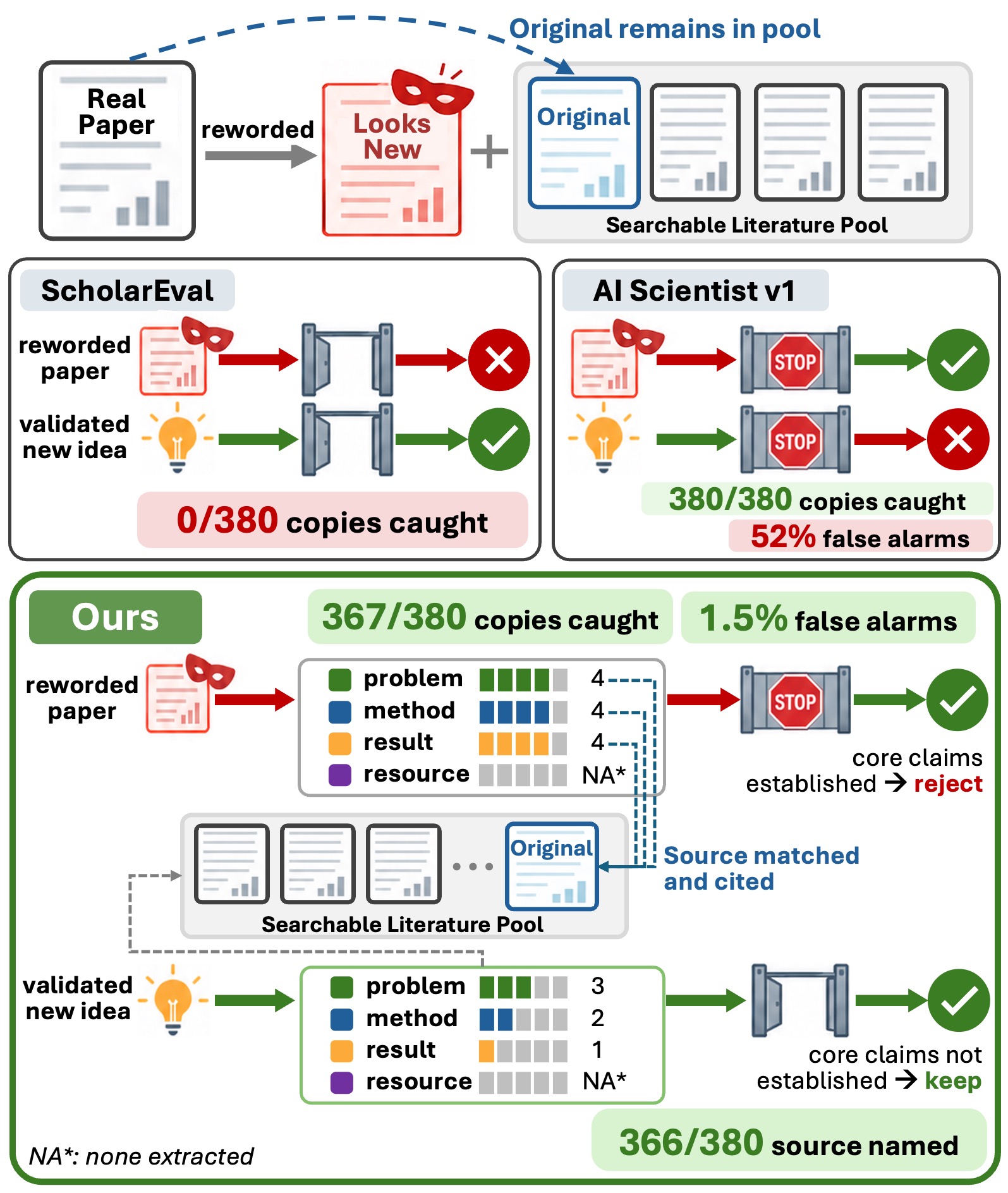}
\caption{A useful novelty module must catch a non-novel copy while keeping a novel idea. We reword $n=95$ published papers into 380 new-looking proposals (four disguises each) and plant the original papers in the pool. Of these disguises (Table~\ref{tab:catchfa}), \textsc{ClaimNov} correctly rejects 367 and names the source in 366, and it rejects only 1.5\% of the 400 later-realized ideas. We show ScholarEval at its default rule (best threshold in Appendix~\ref{app:robust}).}
\label{fig:teaser}
\end{figure}

The development of AI scientific discovery systems is an active, fast-moving research field \citep{lu2026endtoend,yamada2025scientistv2,gottweis2026accelerating,schmidgall2025agent}, and such systems have begun to produce results that look like genuine, highly exciting discoveries \citep{wang2025unstable, sothanaphan2026resolutionerdhosproblem728, yoon2026art}. This is strong evidence that LLMs \textit{can} be creative, motivating their use in hypothesis generation \citep{wang-etal-2024-scimon,si2025llms,liu2026hypobenchsystematicprincipledbenchmarking, oneill2025sparkssciencehypothesisgeneration} and in end-to-end discovery pipelines that generate, filter, and even execute ideas autonomously or semi-autonomously \citep{li-etal-2025-chain-ideas,baek-etal-2025-researchagent,weng2025deepscientistadvancingfrontierpushingscientific,ghareeb2025robinmultiagentautomatingscientific,tang2025airesearcherautonomousscientificinnovation}.

One important facet of creativity, however, is novelty \citep{ismayilzada2025creativepreferenceoptimization, lu2026rethinking}. 
Most AI scientist pipelines employ some kind of novelty check (e.g., searching against Semantic Scholar and using an LLM judge to determine if any results are too close) which in turn decides which generated ideas will go on to receive compute and human attention. Using a novelty check only at the level of a whole proposal has three limitations. (1)~Judging the novelty of a whole proposal is challenging even for human experts; the annotation itself is expensive and annotators often disagree about what counts as novelty \citep{si2025llms}.  (2)~Deployed checks \citep{lu2026endtoend,li-etal-2025-chain-ideas} that reject or pass a research proposal as a whole without considering its parts subtly deviate from the tradition of human science, which is incremental \citep{uzzi2013atypical}. (3)~Whole-proposal checks also make it harder, if not impossible, to understand whether they are actually identifying the right prior work when they reject proposals.
We therefore argue that the right unit to consider is the \textbf{claim}, not the proposal, since most papers, including ones that humans consider highly novel, are not novel in literally every contribution they make \citep{wang2017bias}. With this motivation, we address the limitations of previous work and make the following contributions to the understanding and improvement of novelty checks in AI scientist pipelines:

\begin{itemize}[leftmargin=*,nosep]
    \item We introduce \textbf{\textsc{ClaimNov}}, a fine-grained, interpretable novelty module that treats claims as the basic unit of novelty. 
    \item We propose an \textbf{evaluation framework} for novelty checks in end-to-end AI scientist pipelines which draws upon temporal ground truth from the published record and from reviewer comments at a whole-proposal and claim-level, with four datasets totaling 1{,}060 proposals (\S\ref{sec:materials}).
    \item We audit \textbf{seven existing novelty checks} alongside \textsc{ClaimNov}. The checks that reject the most non-novel work also reject the most work later realized by human researchers, and checks that judge proposals as a whole can only identify claim-level established work by rejecting an entire proposal.
    \item \textsc{ClaimNov} catches 97\% of disguised papers while rejecting only 1.5\% of later realized ideas. It correctly preserves partially novel work, keeping 91 of 100 proposals with a spliced prior contribution against 76 for the strongest existing check, and given the same attribution ranking, it keeps more correctly attributed proposals.
\end{itemize}

\section{Related Work}
\label{sec:related}

\textbf{Traditional measures of scientific novelty.} \citet{zhao2025review} categorize scientific novelty metrics by whether they draw on citation data, textual data, or both. Bibliometric measures use unusual or first-time combinations of cited work \mine{\citep{uzzi2013atypical,wang2017bias,foster2015tradition}} or related ideas such as disruptiveness \citep{wu2019large}; text-based measures consider semantic distance \citep{shibayama2021measuring,jeon2023measuring} or new concepts and new question--method combinations \citep{mishra2016quantifying,luo2022combination}; some measures also combine the two \citep{amplayo2018network}. Traditional measures are often validated against external signals such as expert judgments and later impact \citep{bornmann2019measure,fontana2020new,arts2021natural}. Our evaluation focuses on text-based measures, as does our introduced module \textsc{ClaimNov}, which does not consider bibliometric data.

\textbf{Literature-grounded novelty assessment.} Early on, IR and NLP researchers studied novelty detection at the sentence and document level, as information not covered by previously seen documents \citep{soboroff-harman-2005-novelty,ghosal-etal-2018-novelty,ghosal-etal-2022-novelty}, and patent prior-art searches focused on naming anticipating documents \citep{piroi2011clefip}. LLMs have since enabled approaches that \textit{reason} about the claims of existing work: retrieving prior work to revise an idea \citep{wang-etal-2024-scimon}, judging a paper against retrieved neighbors \citep{lin-etal-2025-evaluating}, reranking retrieved papers by idea-specific facets \citep{shahid-etal-2025-literature}, scoring ideas against a literature set \citep{moussa2025scholareval,qiao2026innoevalresearchideaevaluation}, and comparing atomic content units or contribution claims against full-text evidence \citep{ai-etal-2025-novascore,zhang2026opennovelty,afzal-etal-2026-beyond}. The typed contribution claims used in \textsc{ClaimNov} draw on scientific information extraction \citep{luan-etal-2018-multi,dsouza-etal-2021-semeval}, facet-level views of ideas \citep{radensky2024scideator} and claim verification \citep{wadden-etal-2020-fact}. Benchmarks for these LLM-based systems score them against expert annotations or review text \citep{schopf-farber-2026-idea,wu-etal-2026-novbench}, against future papers \citep{amplayo-etal-2019-evaluating}, or under controlled, counterfactual changes \citep{liu2026axiomatic}. Critiques of LLM novelty judges find that they prefer generated research questions that experts rate lower \citep{sinhahajari2026limits} and that their ratings disagree with hindsight against later papers \citep{jiang2026hindsightevaluatingllmgeneratedresearch}. We specifically evaluate novelty checks as deployed inside AI-scientist pipelines, and also focus on counterfactual changes rather than alignment with explicit human novelty labels, akin to how paraphrase attacks test text detectors \citep{krishna2023paraphrasing}.

\textbf{AI-scientist systems.} Pipelines with a research-ideation stage almost always include a novelty check for generated ideas, although these checks can vary widely in mechanism (Appendix~\ref{app:survey}). Some are hard filters grounded in retrieval, including AI Scientist's agentic search loop \citep{lu2026endtoend}, CoI-Agent's reject-and-regenerate loop \citep{li-etal-2025-chain-ideas}, the retrieval filter of \citet{si2025llms}, and AutoResearchClaw's lexical check \citep{liu2026autoresearchclawselfreinforcingautonomousresearch}. Others are advisory or score-based, with or without explicit retrieval \citep{yamada2025scientistv2,baek-etal-2025-researchagent,gottweis2026accelerating,schmidgall2025agent,weng2025deepscientistadvancingfrontierpushingscientific,weng2025cycleresearcherimprovingautomatedresearch}; still others directly rank generated ideas against one another \citep{ghareeb2025robinmultiagentautomatingscientific,su2025headsbetteroneimproved,tang2025airesearcherautonomousscientificinnovation,keya2025sciideacontextawarescientificideation}.  Work on automatic peer review usually judges full papers, most recently with LLM judges \citep{kang-etal-2018-dataset,wang-etal-2020-reviewrobot,taechoyotin2025remorautomatedpeerreview,chen2025-futureofpeerreview}; we instead focus specifically on novelty.

\section{Novelty Module}
\label{sec:claim-module}

\begin{figure}[!t]
\centering
\includegraphics[width=\columnwidth]{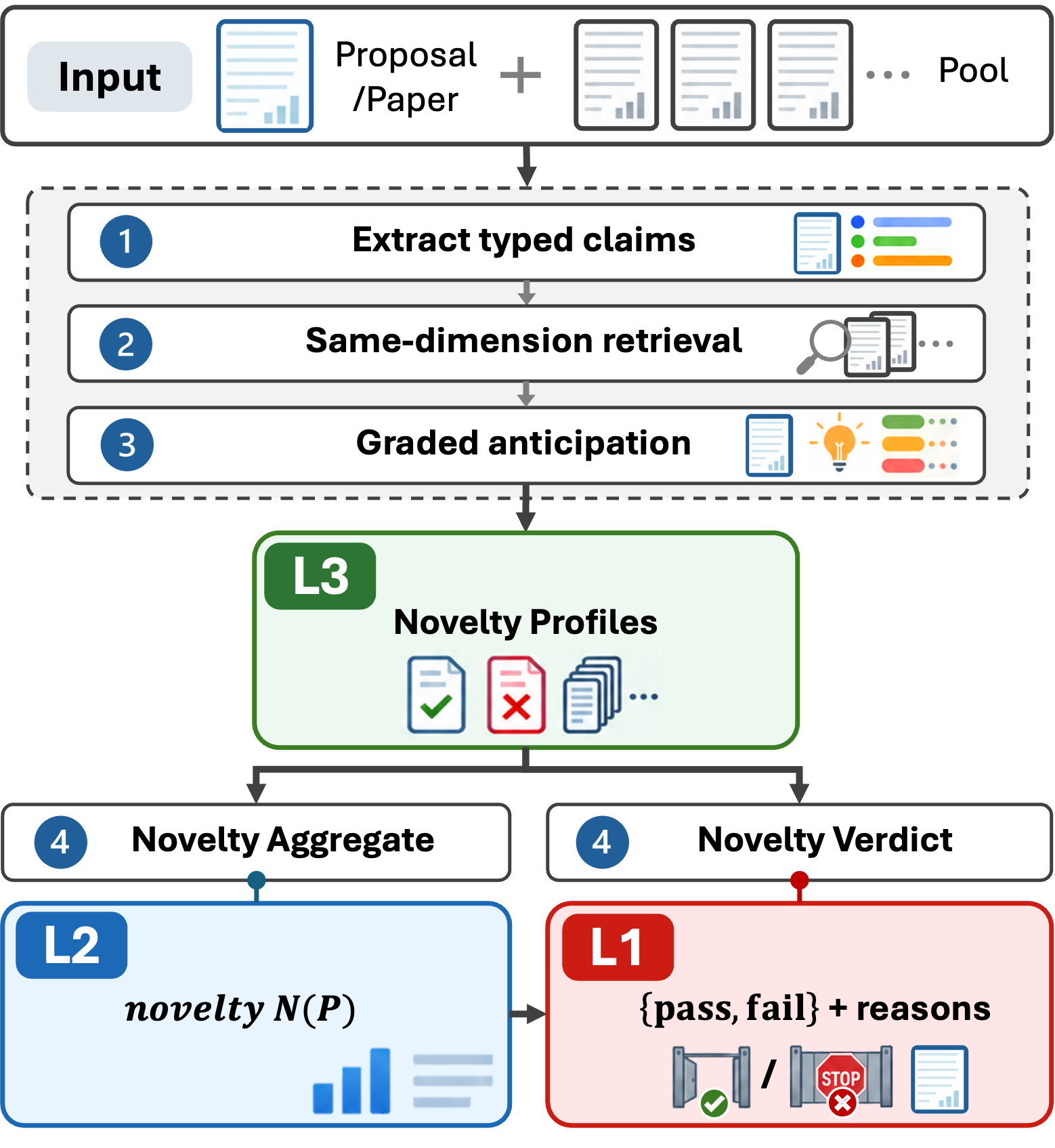}
\caption{\textsc{ClaimNov} extracts typed contribution claims from a proposal $P$, retrieves same-dimension claims from the pool, rates the extent to which each covers $P$'s claims, aggregates these into one novelty score, and gives a full claim-level novelty profile to a verdict judge.}
\label{fig:fig2}
\end{figure}

We introduce \textsc{ClaimNov}, an interpretable, claim-level novelty module for research proposals that can be dropped directly into any AI for scientific discovery pipeline with a novelty check. As shown in Figure \ref{fig:fig2}, \textsc{ClaimNov} extracts typed claims from research proposals, retrieves relevant same-dimension claims from a pool of claims extracted from prior art, scores the entailment per extracted claim, aggregates into a novelty score, and passes the full claim profile to a verdict judge for a binary novel or non-novel decision. For validation experiments, see Appendix~\ref{app:claim-validation}, and for exact prompts, see Appendix~\ref{app:prompts}.

\subsection{Claim Extraction}
\label{sec:extraction}

During the claim extraction stage, \textsc{ClaimNov} takes the title and description from the research proposal $P$ and produces, with an LLM (\judgemodel{} set to temperature 0), a list of contribution claims $C_P = \{c_1,\dots,c_m\}$.

Each claim $c_i$ is typed with a dimension $d_c$ and an attribute-value pair $(c_a, c_v)$. Dimensions are drawn from $\{$Problem, Methodology, Resource, Result$\}$, while an attribute represents the axis along which two works can be compared. For instance, for a Result claim, the attribute may be classification accuracy on CIFAR-10, and the value may be an improvement from 30.1\% to 76.9\%. We aim to extract a small number of claims per proposal (2.4--2.7 on average, the prompt allowing 1--4), based on conventions in AI research. We process each paper in the reference pool $\Pi$ through this stage once, using its abstract as the description text, to produce pool claims $C_\Pi$.

\subsection{Claim Retrieval}
\label{sec:retrieval}

During the claim retrieval stage, \textsc{ClaimNov} takes a claim $c$ from the previous stage's list of claims, embeds its text together with its paper's title with \embmodel{}, and produces a set of relevant pool claims $R(c)$ for comparison to that claim.

To obtain such a set, we retrieve the $k$-nearest (by cosine similarity) pool claims of the same dimension, keeping at most one claim per pool paper. We determine $k$ by pool size, taking $k{=}10$ as the minimum based on an independent retrieval validation (see Appendix~\ref{app:claim-validation}), and for larger pools we use the heuristic $k = 0.2\, n_d$, where $n_d$ is the number of pool papers with a claim in that dimension, capped at 40. This gives $k{=}10$ for the disguise pools and $k \approx 35$ for the anchor pools (\S\ref{sec:materials}).

\subsection{Claim Entailment}
\label{sec:entailment}

During the claim entailment stage, \textsc{ClaimNov} judges the entailment of every pair $(c, c')$ for $c' \in R(c)$ with an LLM (\judgemodel{} at temperature 0), giving titles and abstracts of pool papers as context.

The judge rates the extent to which the pool claim entails or covers the proposal claim on an ordinal scale $\ell(c,c') \in \{1,\dots,5\}$: 1 unrelated, 2 shared topic only, 3 partial coverage with important details differing, 4 strong coverage with minor details missing, and 5 full coverage of the claim's substantive content. We map these levels onto the unit interval by taking $e(c,c') = (\ell(c,c')-1)/4 \in [0,1]$.

\subsection{Novelty Aggregation}
\label{sec:aggregation}

During the novelty aggregation stage, we first take the sorted $e_i$s from the previous stage and combine them into one coverage score $a(c) \in [0,1]$ for each claim as a rank-discounted mean:

\begin{equation}
a(c) \;=\; \frac{\sum_{i=1}^{k_c} \alpha^{i-1} e_i}
                {\sum_{i=1}^{k_c} \alpha^{i-1}},
\qquad \nu(c) \;=\; 1 - a(c),
\label{eq:claim-novelty}
\end{equation}

Here, $\alpha{=}0.5$, which balances the importance of the strongest match against the weaker ones, and $k_c = \min(k, |R(c)|)$, which prevents a claim with very few relevant pool claims from having an inflated score. 

We then combine claim scores into a proposal-level score using dimension weights $w_d$ (Problem 0.4, Result 0.3, Methodology 0.2, Resource 0.1) and per-claim importance weights $\mathrm{imp}(c)$, which are uniform for proposals and set to each claim's share of support passages when scoring a full paper. The dimension weights encode a prior about contribution types in AI research; they can be tuned to adapt \textsc{ClaimNov} to other fields.

\begin{equation}
\begin{aligned}
A(P) &= \textstyle\sum_{c \in C_P} w_{d_c}\, \mathrm{imp}(c)\, a(c), \\
A_{\max}(P) &= \textstyle\sum_{c \in C_P} w_{d_c}\, \mathrm{imp}(c), \\
N(P) &= 1 - A(P)/A_{\max}(P).
\end{aligned}
\label{eq:paper-novelty}
\end{equation}

$A_{\max}(P)$ is the coverage if every claim were fully covered. $N(P)$ summarizes the novelty profile in one number; the full profile is provided as context to the novelty verdict stage below.

\subsection{Novelty Verdict}
\label{sec:verdict}

Finally, the novelty verdict stage of \textsc{ClaimNov} assembles previous evidence produced by the module into a profile $\Phi(P)$ which is evaluated by a stronger judge (\texttt{claude-sonnet-4-6}). $\Phi(P)$ lists the proposal's extracted claims with matched prior claims, their sources (and their abstracts), and the entailment levels.

From the judge, we receive a verdict of pass or fail. If the proposal fails, \textsc{ClaimNov} also outputs the prior art paper $p^\ast \in \Pi$, its subset of claims $S \subseteq C_P$, and a one- or two-sentence reason. For the exact prompt, please see Appendix~\ref{app:prompts}.

\section{Experimental Setup}
\label{sec:setup}

We explain the shared inputs and outputs for each check, which existing checks we evaluate, the judge models and their temporal cutoffs, the test data we collect, and the metrics we define.

\subsection{Evaluation Contract}
\label{sec:contract}

Because each of the AI scientist pipelines and standalone scientific novelty evaluations we survey (Appendix~\ref{app:survey}) uses slightly different schema, literature search, and verdict formats, we standardize the evaluation contract to allow for better comparison.

We pass as input to each check a \textit{research proposal} consisting of a title plus a 150--400 word description as well as a \textit{reference pool}, which is a list of prior art papers given as $\{\text{id}, \text{title}, \text{abstract}\}$. Checks using search are sandboxed. For more details on how we adapt proposals to native input formats, see Table~\ref{tab:survey-adapters}. For output, we consider three layers: L1, the verdict $\textit{verdict} \in \{\textit{pass}, \textit{fail}\}$ with reasons; L2, a novelty score $N(P) \in [0,1]$; and L3, a novelty profile, every claim with its matched prior claims, their entailment levels, and the papers they come from. For \textsc{ClaimNov}, L3 determines both L2 and L1 (\S\ref{sec:aggregation}, \S\ref{sec:verdict}). The existing checks we evaluate only emit L1.

\subsection{Evaluated Systems}
\label{sec:systems}

We consider four kinds of novelty check. (1) Novelty checks that are a clear-cut component in some end-to-end AI-scientist pipeline: AI Scientist~v1 (\fnvone{}), CoI-Agent (\fncoi{}), Si et al.\ (\fnsi{}), and ARC's lexical filter. (2) Novelty checks we have ``extracted'' from pipelines that make some kind of novelty judgment but which are not clearly plug-and-play in the same way: ResearchAgent's Originality rating (a proposal passes at a rating of 4 or 5 out of 5), ScholarEval's pairwise contribution scoring (a proposal passes when its mean score against the top-8 pool papers is positive), and AI-Scientist-v2's ideation reflection. The v2 ideation reflection has no final verdict, so we append a final verdict to the prompt. (3) A claim-level reference baseline after the style of OpenNovelty \citep{zhang2026opennovelty}, which judges 1--3 extracted contributions against their BM25 top-10 and rejects when every contribution is refuted. (4) \textsc{ClaimNov} as described in \S\ref{sec:claim-module} and its ablations. Appendix~\ref{app:survey} gives our selection criteria and the full survey of 17 pipelines. Searches embedded in novelty checks are replaced by dense retrieval over the pool; see Appendix~\ref{app:adapters} for details.

\subsection{Judge Models and Temporal Validity}
\label{sec:models}

Because six of the seven novelty checks we evaluate use an LLM as a judge at some point (and \textsc{ClaimNov} also does so), we fix a cutoff $T=$ March 2026 and exclusively evaluate models with earlier cutoffs. This mitigates the risk of models recognizing post-$T$ papers from training data. We run each check with Sonnet~4.6 (training cutoff January 2026), which is the same model \textsc{ClaimNov} uses as the final verdict judge (for earlier, lightweight stages, \textsc{ClaimNov} uses \judgemodel{}). As we explain in more detail in \S\ref{sec:materials}, we use Sonnet~4.6 to generate ideas, disguise existing papers, and rewrite disputed papers as research proposals, and Opus~4.6 (training cutoff August 2025) as the labeler.

\subsection{Data Collection}
\label{sec:materials}

We collect data and establish ground truth for two levels of non-novelty. In both cases, we aim to collect data that falls after our temporal cutoff of $T=$ March 2026, to avoid LLM judge awareness.

\paragraph{Already-done ideas ($n{=}400$, whole-proposal level).}
For this data, novelty checks should correctly reject a disguised proposal if the source paper is in the pool, but should \textit{not} reject when the source is removed (making the proposal novel relative to the pool). We collect $n=100$ ICML 2026 spotlight papers, filtering for papers where the v1 arXiv appeared \textit{after} $T$, and disguise each paper as a new research proposal at four increasingly challenging to spot levels of disguise. Rung 1 swaps the voice, rung 2 paraphrases, rung 3 reframes the paper with a different motivation, reordered contributions, and an unnamed method, and rung 4 restructures it adversarially from a different theoretical starting point with generalized numbers. Because rungs 3 and 4 may depart significantly from the original paper, we accept them only when a verifier (Sonnet~4.6) confirms that the core contributions are the same. To ensure the disguise is challenging, we require that at most 2\% of the disguise's 6-grams appear in the source, regenerating up to three times; for more details and prompts, see Appendix~\ref{app:disguise-prompts}. We then form a reference pool from the source plus up to 19 pre-$T$ papers closest to the proposal by embedding, which we take from the source's reference list and Semantic Scholar recommendations. We drop five sources whose pools contain only the source, leaving 380 disguises (all 400 in Appendix~\ref{app:robust}).

\paragraph{Claim-disputed ideas ($n{=}160$, claim level).}
For this data, novelty checks should correctly find a prior work $B$ that establishes a specific claim. We use a regex prefilter and gpt-5-nano classification to mine 8.3k ICLR 2022--23 reviews in which a reviewer identified a paper $A$'s contribution as already established by some prior paper $B$. We then rewrite each $A$ in proposal voice with Sonnet~4.6 and the rung-1 prompt provided in Appendix~\ref{app:disguise-prompts} and plant $B$ among 19 pre-$T$ papers to serve as distractors. We also record the final ICLR decision (62 accepted, 96 rejected, 2 unknown) as an additional axis of analysis of novelty check performance. Of $n=200$ pairs of papers $(A,B)$, we keep $n=160$ with usable pools, and we annotate a sample for genuine disputes (see Appendix~\ref{app:labels}). Unlike the other sets, these papers predate $T$ and may appear in a judge's training data. In this case the ground truth is reviewer attribution (and not some post-$T$ event), so we consider this acceptable.

\paragraph{Later-realized ideas ($n{=}400$, whole-proposal level).}
For this data, novelty checks should correctly \textit{pass} the proposal, because the pool does not contain the idea and researchers went on to realize some of its contributions. We collect $n{=}30$ anchor papers, where each anchor paper was published in NeurIPS 2025 proceedings and has at least 50 citing papers that fall after $T$. We exclude benchmarks, datasets, and surveys (which may be cited primarily for use), then generate new proposals based on the remaining anchor abstracts. We construct a relevant works pool (median 174 papers, capped at 400 papers) for each anchor from its reference list plus any citing papers that come before $T$, as well as a follow-up pool $F$ of the citing papers that come after $T$ (median 124 papers). We use each anchor paper to generate $n{=}100$ ideas from Sonnet~4.6, producing $n{=}3000$ proposals total. For each of these proposals, Opus~4.6 writes a contribution list, then labels each contribution as established or not by the 8 papers in the pool nearest to the idea (by \texttt{text-embedding-3-small} over title and abstract) and, separately, by the 8 nearest papers in $F$. We identify later-realized ideas as proposals with at least one contribution established in $F$ and none established in the pool. From the 407 proposals of this kind generated from 26 of the 30 anchors, we draw 400 at random to constitute this dataset. We provide labeling prompts in Appendix~\ref{app:label-prompts}.

\paragraph{Hybrid proposals ($n{=}100$, claim level).}
For this data, novelty checks should correctly pass the proposal \textit{and} correctly name the paper whose contribution it contains. We take $n=100$ later-realized (and thus assumed novel) ideas from the previous dataset and, for each, a pool paper $B$ which was moderately similar (ranked 5--15 by cosine similarity), then ask Sonnet~4.6 to rewrite these ideas so that $B$'s core contribution is also included. Sonnet~4.6 verifies that the splice preserved the original contributions while adding the new one.

\subsection{Metrics}
\label{sec:metrics}
We evaluate based on two levels of granularity -- the whole-proposal level and the claim level -- motivated by the idea that we want to reject proposals that are wholly non-novel, but not proposals that simply contain some existing claims.

For the whole-proposal level, we evaluate checks' ability to \textbf{catch} non-novel ideas as well as their \textbf{false alarm rate}, how often they reject ideas that are later realized and validated by the human researcher community. We measure false alarms both on the later-realized ideas (FA) and, as a second set of assumed novel ideas, on the disguises with their source removed from the pool (Src.\ out). From these, we compute informedness $\textbf{J} = \text{catch} - \text{FA}$ and the \textbf{pool-flip} $\bm{\Delta} = \text{catch} - \text{Src.\ out}$, or the drop in rejection rate when the disguised proposal's source paper is removed from the pool.

For both whole-proposal and claim level, we evaluate checks' ability to \textbf{attribute} claims to prior art. For whole-proposal, we observe whether checks name the source paper for disguises, and for claim level, we report the share of the $n{=}160$ claim-disputed proposals for which a check's top-ranked paper is the prior art $B$. We also report rejection rates on this set by ICLR decision, with the motivation that one disputed claim does not a non-novel proposal make.

\section{Results}
\label{sec:results}

We find that \textsc{ClaimNov} is both sensitive and specific, catching disguised copies of published work while keeping ideas that researchers went on to publish (\S\ref{sec:catchfa}). Because its verdict rests on \textit{claim-level} evidence, we also find that \textsc{ClaimNov} preserves proposals in which a single contribution is established, so more correctly attributed proposals can reach the full AI-scientist pipeline (\S\ref{sec:claim}, \S\ref{sec:hybrid}). We report full results in Tables~\ref{tab:catchfa}--\ref{tab:hybrid}, with 95\% intervals and paired exact McNemar tests in Appendix~\ref{app:stats}.

\subsection{Whole-Proposal Evaluation}
\label{sec:catchfa}

\begin{table*}[t]
\centering
\small
\setlength{\tabcolsep}{6.5pt}
\begin{tabular}{llrrrr>{\columncolor{gray!10}}r>{\columncolor{gray!10}}r}
\toprule
 & & \multicolumn{4}{c}{\textsc{rates}} & \multicolumn{2}{c}{\textbf{\textsc{net scores}}} \\
\cmidrule(lr){3-6}\cmidrule(l){7-8}
\rowcolor{gray!15} & \textbf{Check} & \textbf{Caught}$\uparrow$ & \textbf{Named}$\uparrow$ & \textbf{Src.\ out}$\downarrow$ & \textbf{FA}$\downarrow$ & $\bm{J}\uparrow$ & $\bm{\Delta}\uparrow$ \\
\midrule
\multirow{4}{*}{\rotatebox{90}{\scriptsize Deployed}}
 & AI Scientist v1 & \textbf{380} & 292 & 91 & 51.8\% & +0.48 & +0.76 \\
 & CoI-Agent & 368 & 362 & 31 & 3.8\% & \underline{+0.93} & +0.89 \\
 & Si et al. & \textbf{380} & \textbf{379} & 359 & 96.0\% & +0.04 & +0.06 \\
 & ARC & 85 & 85 & \textbf{0} & \textbf{0.0\%} & +0.22 & +0.22 \\
\midrule
\multirow{3}{*}{\rotatebox{90}{\scriptsize Extract.}}
 & ResearchAgent$^\dagger$ & 338 & --- & 260 & 98.5\% & -0.10 & +0.21 \\
 & ScholarEval$^\dagger$ & 0 & --- & \textbf{0} & \textbf{0.0\%} & +0.00 & +0.00 \\
 & AI-Scientist-v2$^\dagger$ & 362 & --- & 117 & 39.2\% & +0.56 & +0.64 \\
\midrule
 & OpenNovelty-style & \underline{379} & \textbf{379} & 38 & 11.5\% & +0.88 & \underline{+0.90} \\
\midrule
\rowcolor{gray!25} & \textbf{\textsc{ClaimNov} (ours)} & 367 & \underline{366} & \underline{18} & \underline{1.5\%} & \textbf{+0.95} & \textbf{+0.92} \\
 & \;\emph{-- w/o claim profile (judge over top-8 abstracts)} & 373 & 367 & 57 & 11.8\% & +0.86 & +0.83 \\
 & \;\emph{-- w/o verdict judge (graded check)} & 349 & 345 & 20 & 1.8\% & +0.90 & +0.87 \\
 & \;\emph{-- per-claim verdict rule} & 380 & 360 & 197 & 56.8\% & +0.43 & +0.48 \\
 & \;\emph{-- fixed retrieval depth ($k{=}10$)} & 367 & 366 & 18 & 2.0\% & +0.95 & +0.92 \\
\bottomrule
\end{tabular}
\caption{Evaluation results on whole-proposal data for all checks, where every check is run with Sonnet~4.6 as its judge (excluding ARC, which is lexical). Caught and Src.\ out count rejections of the $n=380$ disguised proposals with the source in and out of the pool; Named counts rejections that name the source; FA is on the 400 later-realized ideas. $J$ and $\Delta$ are defined in \S\ref{sec:metrics}; see Table~\ref{tab:ci} for 95\% cluster-bootstrap intervals. $^\dagger$Decision rule set by us (Appendix~\ref{app:adapters}); best in-sample thresholds are in Appendix~\ref{app:robust}. For AI Scientist~v1, ``Named'' counts any of up to three papers. Note here the fixed-depth row is equivalent to the main row because the scaled depth is also 10 on 20-paper disguise pools. Bold: best, underline: second best, ablation rows excluded.}
\label{tab:catchfa}
\end{table*}

On whole proposals (Table~\ref{tab:catchfa}), all checks except ARC and ScholarEval catch at least 338 of 380 disguises, but most checks \textit{continue} to reject after the source is removed. For instance, Si et al.\ catches all 380 but also rejects 359 when the source is no longer present in the pool. All checks except \textsc{ClaimNov} and CoI-Agent also reject a large share of ideas the field went on to publish. Si et al.\ rejects 96\%, ResearchAgent 98.5\%, AI Scientist~v1 52\%, and AI-Scientist-v2 39\%. ARC and ScholarEval reject 0.0\%, but do so by rejecting almost nothing.

\textsc{ClaimNov} and CoI-Agent are clear outliers in their ability to balance the two. Both catch almost every disguise (367 and 368 of 380, respectively) while rejecting few later-realized ideas (1.5\% and 3.8\%). \textsc{ClaimNov} has the highest $J$ ($+0.95$) and $\Delta$ ($+0.92$), whereas the next best pipeline checks after CoI-Agent are at $+0.56$ and $+0.76$. The per-contribution OpenNovelty-style baseline does catch more disguises (379, $p{<}.001$) but false-alarms on 11.5\% of later-realized ideas and rejects 38 disguises with the source removed ($p{=}.001$ against \textsc{ClaimNov}). While clustering tests by source and anchor shows \textsc{ClaimNov} and CoI-Agent are statistically level, \textsc{ClaimNov}'s verdict call costs a quarter of CoI-Agent's check call with the same judge (\$0.010 against \$0.039), with entailment cost growing with pool size (Table~\ref{tab:cost}, Appendix~\ref{app:loop}).

The \textsc{ClaimNov} ablations (Appendix~\ref{app:ablations}) show the value of each stage of the module. Without the claim-level novelty profile, the verdict judge reads the nearest abstracts instead, and its false-alarm rate rises from 1.5\% to 11.8\% as a result. When we instead use a per-claim verdict rule, which fails a proposal if any central claim appears entailed, we catch all 380 disguised proposals, but the false-alarm rate rises to 56.8\% and we wrongly reject 197 of the 380 proposals whose source has been removed. This reinforces the idea that (1) claims are the appropriate unit for understanding a research proposal and (2) a research proposal is \textit{not} one large claim; its claims should be considered separately. On generated ideas that a whole-idea labeler matched to one pool paper, \textsc{ClaimNov} rejects 18\%, against 55\% for the judge without the claim-level novelty profile (Appendix~\ref{app:natural}). These ideas are noisier than disguises, and we audit a sample of them by hand, together with later-realized ideas, in Appendix~\ref{app:audit}. Results with other verdict judges are given in Appendix~\ref{app:robust}.

\subsection{Claim-Level Prior Art}
\label{sec:claim}

\begin{table}[t]
\centering
\small
\setlength{\tabcolsep}{1.5pt}
\resizebox{\columnwidth}{!}{\begin{tabular}{lrrrr}
\toprule
\rowcolor{gray!15}\textbf{Check} & \textbf{Rej.\ acc.}$\downarrow$ & \textbf{Rej.\ rej.} & \textbf{Names $B$}$\uparrow$ & \textbf{on pass}$\uparrow$ \\
\midrule
AI Scientist v1 & 53\% & 40\% & 20\% & 0\% \\
CoI-Agent & 10\% & 11\% & 6\% & 0\% \\
Si et al. & 81\% & 78\% & \textbf{47\%} & 0\% \\
ARC & 3\% & 0\% & 0\% & 0\% \\
ResearchAgent & 47\% & 66\% & --- & --- \\
ScholarEval & 0\% & 0\% & --- & --- \\
AI-Sci.-v2 & 16\% & 14\% & --- & --- \\
\midrule
OpenNov.-style & 8\% & 7\% & 17\% & 12\% \\
\midrule
\rowcolor{gray!25}\textbf{\textsc{ClaimNov} (ours)} & 3\% & 3\% & 21\% & 20\% \\
\;\emph{-- w/o claim profile} & 13\% & 9\% & 15\% & 7\% \\
\;\emph{-- w/o verdict judge} & 6\% & 9\% & 21\% & 20\% \\
\;\emph{-- per-claim rule} & 21\% & 20\% & 21\% & 15\% \\
\midrule
\multicolumn{5}{l}{\emph{Verdict + shared ranking}} \\
CoI + dense top-1 & 10\% & 11\% & \underline{37\%} & \underline{31\%} \\
CoI + BM25 top-1 & 10\% & 11\% & 26\% & 23\% \\
\rowcolor{gray!25}\textbf{\textsc{ClaimNov} + dense (ours)} & 3\% & 3\% & \underline{37\%} & \textbf{36\%} \\
\rowcolor{gray!25}\textbf{\textsc{ClaimNov} + BM25 (ours)} & 3\% & 3\% & 26\% & 24\% \\
\bottomrule
\end{tabular}}
\caption{Claim-level prior art ($n{=}160$, Sonnet~4.6 judge). Rej.\ acc./Rej.\ rej.: rejection rate on the 62 ICLR-accepted and 96 ICLR-rejected proposals (lower is better only for accepted proposals). Names $B$: share whose top-ranked paper is the reviewer's $B$. On pass: same, counting only passed proposals. Note most existing checks name a paper only when rejecting. The bottom rows pair a verdict with the pool paper nearest to the whole proposal by dense embedding or BM25 to separate final verdict and attribution. --- means no paper returned. Bold: best, underline: second.}
\label{tab:claim}
\end{table}

We evaluate \textsc{ClaimNov} and the seven existing checks on claim-level data (Table~\ref{tab:claim}) and find that \textsc{ClaimNov} correctly treats a disputed claim as a claim-level finding, and not a reason to reject an entire research proposal. It rejects only 3\% of accepted ICLR proposals, the lowest rate of any check that catches disguises (10\% for CoI-Agent and 47--81\% for Si et al., AI Scientist~v1, and ResearchAgent). All of the whole-proposal checks we evaluate name prior art only alongside a rejection, so Si et al.\ names $B$ for 47\% by rejecting about 80\% of proposals, and CoI-Agent passes 89\%, meaning it can name $B$ for only the remaining 11\%.

Considering this design choice in existing checks, we separate attribution from the verdict by pairing each verdict with the \textit{same} attribution ranking. \textsc{ClaimNov}'s top-rated profile paper is $B$ for 21\% of proposals, while the pool paper nearest to the whole proposal by dense embedding is $B$ for 37\%. Intuitively, this is because the $B$ named by a reviewer is often closest to the proposal as a whole, which embedding similarity captures well, whereas \textsc{ClaimNov} ranks papers by how fully they cover a \textit{single} extracted claim. Extraction failures further exacerbate this issue (Appendix~\ref{app:attribution}). We therefore pass the dense ranking to every verdict, and \textsc{ClaimNov} then passes the proposal and names $B$ for 36\% of proposals, against 31\% for CoI-Agent ($p{=}.016$).  Given the same attribution, \textsc{ClaimNov}'s verdict keeps more proposals building on established work.

\subsection{Hybrid Proposals}
\label{sec:hybrid}

\begin{table}[t]
\centering
\small
\setlength{\tabcolsep}{4pt}
\resizebox{\columnwidth}{!}{\begin{tabular}{lrrr}
\toprule
\rowcolor{gray!15}\textbf{Check} & \textbf{Passes}$\uparrow$ & \textbf{Names $B$}$\uparrow$ & \textbf{on pass}$\uparrow$ \\
\midrule
AI Scientist v1 & 18 & 23 & 0 \\
CoI-Agent & 76 & 11 & 0 \\
Si et al. & 0 & 63 & 0 \\
ARC & \textbf{100} & 0 & 0 \\
ResearchAgent & 2 & --- & --- \\
ScholarEval & \textbf{100} & --- & --- \\
AI-Scientist-v2 & 29 & --- & --- \\
\midrule
OpenNov.-style & 83 & 68 & \underline{64} \\
\midrule
\rowcolor{gray!25}\textbf{\textsc{ClaimNov} (ours)} & \underline{91} & 25 & 18 \\
\;\emph{-- w/o claim profile} & 68 & 42 & 14 \\
\midrule
\multicolumn{4}{l}{\emph{Verdict + shared ranking}} \\
CoI + dense top-1 & 76 & 7 & 6 \\
CoI + BM25 top-1 & 76 & \textbf{74} & 53 \\
\rowcolor{gray!25}\textbf{\textsc{ClaimNov} + dense (ours)} & \underline{91} & 7 & 7 \\
\rowcolor{gray!25}\textbf{\textsc{ClaimNov} + BM25 (ours)} & \underline{91} & \textbf{74} & \textbf{67} \\
\bottomrule
\end{tabular}}
\caption{Hybrid proposals ($n{=}100$, Sonnet~4.6 judge). Names $B$: the top-ranked paper is the spliced paper $B$ (as in Table~\ref{tab:claim}). On pass: $B$ is named and the proposal passes. ---: no paper returned. Bold: best, underline: second.}
\label{tab:hybrid}
\end{table}

Evaluating \textsc{ClaimNov} alongside the existing checks on the hybrid proposals (Table~\ref{tab:hybrid}) allows us to more clearly distinguish \textsc{ClaimNov}'s advantages. We fix the correct behavior by construction, as each proposal is a later-realized novel idea which should \textit{not} be rejected, and we control proper attribution by splicing in a contribution from a known paper $B$. The other checks continue to behave roughly according to established traits. ARC and ScholarEval pass all 100 and name nothing, Si et al.\ names $B$ for 63 but rejects all 100, and ResearchAgent rejects 98. CoI-Agent, the closest check on whole proposals, passes 76 and names $B$ for only 11 of the 100, all among the 24 it rejects.

\textsc{ClaimNov}, on the other hand, passes 91 of 100 ($p{=}.004$ against CoI-Agent). Because a hybrid states $B$'s contribution explicitly, BM25 over the whole proposal ranks $B$ first for 74. Given this shared ranking (similarly to the isolation we perform in \S\ref{sec:claim}), \textsc{ClaimNov} keeps 67 hybrids with $B$ named, against 53 for CoI-Agent ($p{=}.004$). The OpenNovelty-style baseline keeps 64, but this is at the cost of 11.5\% false alarms on later-realized ideas (Table~\ref{tab:catchfa}). When \textsc{ClaimNov} successfully extracts the spliced claim as its own claim (as opposed to missing it or absorbing it into a different claim), the entailment judge rates $B$ at level 4 or 5, and the verdict stage properly reads one established claim among several novel ones as a proposal that builds on prior work. We consider these ratings reliable, as 13 of the 15 papers rated at level 4 or above are $B$, and this entailment level is reached by only 9 of 400 later-realized ideas (Appendix~\ref{app:stats}).

\section{Conclusion}
\label{sec:conclusion}

We establish an evaluation framework with temporal ground truth at both the proposal and the claim level, and use it to audit seven existing novelty checks. We find that the checks which reject the most non-novel work also tend to reject the most work the field went on to publish, and observe that checks built to judge proposals as a whole name claim-level prior art only when they reject. We therefore introduce \textsc{ClaimNov}, a drop-in claim-level novelty module, and show it rejects non-novel ideas, keeps novel and partially novel ones, and, given the same attribution ranking, keeps more correctly attributed proposals. Beyond filtering, a specific claim-level check like \textsc{ClaimNov} could serve as a reward when training AI-scientist agents, since it does not penalize the incremental follow-ups most science consists of, and its profile tells the agent which of its contributions are new. Do AI scientists know what's new? Based on current novelty checks, they largely do not, rejecting much of what the field goes on to publish or practically nothing at all. However, \textsc{ClaimNov} enables scientific novelty to be judged \textit{claim by claim} against specific prior papers, thus potentially making AI scientists much better judges of novelty.

\section*{Limitations}

Our work employs several heuristics in how we consider sets of proposals or papers to be novel or non-novel, and this is not as clean as if we had a ``true'' dataset. We also focus exclusively on AI research. In practice, this is where most AI-scientist work also concentrates, so this aligns well, but if we want AI-scientist pipelines to be more widely applicable, then assumptions like the ones we make about dimension weights (problem contributions being more valuable than other contributions, for instance) might not generalize and would require further tuning for different fields, thus impeding adoption. We also consider only English-language research, which can further skew generalization. 

Because our pools are fixed lists of prior work rather than open search, consisting of 20 papers for disguises and disputed pairs and up to 400 for generated ideas, we have limited ability to evaluate the \textit{retrieval} of novelty checks. Additionally, we identify later-realized ideas with an LLM labeler, although we do cross-check this with a different model family labeler and with a small-scale human audit in Appendix~\ref{app:audit}. Our disguised proposals and generated ideas are generated by the same model (Sonnet~4.6) that ultimately judges, which is a known source of possible bias for LLMs-as-judges \citep{zheng2023judging,panickssery2024llm}.

The temporal construction of our datasets also decays, since running the same analysis with more recent LLM judges (with training cutoffs post-$T$) will require all materials to be rebuilt at a later $T$. Although this is simple mechanically and our scripts support it, this still takes time and money. Finally, one important aspect of novelty modules in AI-scientist pipelines is how they affect in-the-loop generation: when the module is actively rejecting and causing regeneration of proposals, how does that affect the final committed set of ideas, and what are the compute and token costs involved? We leave this question to future work because (1) it is highly computationally expensive compared to the static question we examine and (2) establishing ground truth for ideas produced in response to a rejection is harder still and may require extensive human annotation, for which we have limited resources. We do run a small pilot (Appendix~\ref{app:loop}) and report the rejection and cost dynamics of the loop.

\section*{Ethical Considerations}
Our aim with this work is to encourage the development of scientific novelty modules in AI scientist pipelines which are better aligned with human understandings of novelty, which we believe is important to the science produced by these pipelines. We recognize that development of AI for scientific discovery capabilities has several concerning implications for current review processes and for the human experience of science in general. \textsc{ClaimNov}, like any automated judge, should be used carefully.

We do not release any intellectual property directly; we release no direct paper text, only pointers and code for reproduction of our results, and we do not release ICLR review text, but rather the code that re-mines the claim disputes from OpenReview. We follow the terms of use of both arXiv and OpenReview licenses. Additionally, while we use human annotation to validate some LLM labels, the annotation consists only of labels on published papers and generated ideas, and we collect no survey responses or personal information. It is not human-subjects research and falls outside IRB review.

\section*{Acknowledgments}
We thank Lucy Zhou for her work on an earlier version of this project. Miri Liu was supported by the Amazon AI PhD Fellowship and by the U.S. Department of Energy, Office of Science, Office of Advanced Scientific Computing Research, Department of Energy Computational Science Graduate Fellowship under Award Number DE-SC0026617. This work is supported in part by the National Science Foundation under Grants 2229612 and 2433308. Any opinions, findings, and conclusions or recommendations expressed in this material are those of the authors and do not necessarily reflect the views of the United States Government or any agency thereof.

\bibliography{custom}

\appendix

\FloatBarrier
\section{Pipeline Survey}
\label{app:survey}
Table~\ref{tab:survey-embedded} surveys the novelty judgments embedded in AI-scientist pipelines, and Table~\ref{tab:survey-standalone} lists standalone novelty judges. Table~\ref{tab:survey-committed} shows what each pipeline passes downstream after its check, which shapes the evaluation contract (\S\ref{sec:contract}). Table~\ref{tab:survey-select} gives the systems we include and exclude. We include a system when (1) its code is available, (2) it has a separable judgment call, (3) that judgment takes a research proposal as input, (4) we can control the reference pool or sandbox its retrieval, and (5) its judge can be replaced with one whose training cutoff predates $T$. Table~\ref{tab:survey-adapters} shows how the shared proposal is passed to each included check, and Table~\ref{tab:cost} gives the measured cost of the final judge calls.

\begin{table*}[!htbp]
\centering
\scriptsize
\setlength{\tabcolsep}{3pt}
\renewcommand{\arraystretch}{1.08}
\begin{tabular}{>{\raggedright\arraybackslash}p{1.65cm}>{\raggedright\arraybackslash}p{2.4cm}>{\raggedright\arraybackslash}p{2.75cm}>{\raggedright\arraybackslash}p{2.7cm}>{\raggedright\arraybackslash}p{3.1cm}>{\raggedright\arraybackslash}p{2.3cm}}
\toprule
\rowcolor{gray!15}\textbf{System} & \textbf{Where judgment sits} & \textbf{Input artifact $I$} & \textbf{Pool at judgment time} & \textbf{Output $O$ and enforcement} & \textbf{Pool controllable?} \\
\midrule
AI Scientist v1 \citep{lu2026endtoend} & Hard post-ideation filter (\fnvone) & Idea dict: Name, Title, Experiment (2--4 sentences), 1--10 self-ratings; task description and template code & Agentic loop, $\le$10 rounds $\times$ 10 Semantic Scholar hits (title, abstract, venue, year, citations) & \texttt{novel}: bool parsed from a literal ``Decision made''; non-novel ideas discarded; no decision after 10 rounds $\Rightarrow$ not novel & Yes: one \texttt{search\_for\_papers()}; replace \fnvone{} wholesale \\
\addlinespace[2pt]
AI Scientist v2 \citep{yamada2025scientistv2} & No separable check; advisory retrieval inside ideation reflections & Search-query strings; idea JSON (Name, Title, Short Hypothesis, Related Work, Abstract $\approx$250w, Experiments) & Same endpoint, top-10, citation-sorted & Nothing machine-readable; lands as Related Work prose; a human picks the idea to run & Yes: \texttt{use\_tool(query)} is the single chokepoint \\
\addlinespace[2pt]
CoI-Agent \citep{li-etal-2025-chain-ideas} & Hard filter plus uncapped regeneration loop & One free-text blob (motivation, novelty, method merged; structure dissolved before the check) & $\le$3 LLM queries $\times$ 5 hits, title+abstract; date cutoff and ban list built in & Binary \texttt{<similar>} plus offending paper and summary, fed back to regeneration; criterion near plagiarism-strength identity; empty retrieval $\Rightarrow$ novel & Yes: \fncoi{} has a clean signature \\
\addlinespace[2pt]
Si et al.\ \citep{si2025llms} & (a) intra-batch embedding dedup; (b) Swiss tournament, novelty excluded; (c) retrieval filter, the real check & (a) title + 5 fields; (c) full proposal JSON (Problem, Existing Methods, Motivation, Proposed Method, Experiment Plan; 1--2 pages) & (a) own ideas only, MiniLM cosine $>0.8$; (c) top-10 abstracts via LLM keyword queries & (a) drop duplicates (4000 $\to$ $\approx$200); (c) per-paper binary ``same problem and same approach?''; any yes rejects, evidence attached & Yes: \fnsi{}, or pre-populate the paper list \\
\addlinespace[2pt]
ResearchAgent \citep{baek-etal-2025-researchagent} & Soft critic in generate $\to$ validate $\to$ refine $\times$3, per artifact (problem, method, experiment) & Artifact plus rationale; grounding context (core paper, references without abstracts, entity store) & No retrieval at review time; frozen jsonl pool from generation & Originality 1--5 Likert plus feedback; steers refinement and best-of-$N$; never rejects & Yes (fully): static pool; swap the originality prompt \\
\addlinespace[2pt]
Agent Laboratory \citep{schmidgall2025agent} & No check; lit review judges relevance; post-hoc reviewer scores the finished paper & Topic string (review); full LaTeX (reviewer) & arXiv top-5 (review); parametric (reviewer) & Originality 1--4 inside a NeurIPS-style review JSON; revision feedback and solver reward; never checks pre-experiment & Yes: \texttt{ArxivSearch} swappable; insert a phase \\
\addlinespace[2pt]
SCI-IDEA \citep{keya2025sciideacontextawarescientificideation} & Mid-pipeline ``Aha'' filter & 1--2 sentence idea, SciBERT-embedded & None; novelty $=1-\max$ cosine to sibling candidates, surprise $=-\log p$ & Continuous scores $\to$ binary flag; only flagged ideas refined & No code released \\
\addlinespace[2pt]
CycleResearcher \citep{weng2025cycleresearcherimprovingautomatedresearch} & Post-hoc whole-paper review; reward for RL loop & Full paper ($\approx$14k tokens) & Parametric (fine-tuned on ICLR reviews) & Review dict (ratings, contribution, decision); no novelty field & No: monolithic fine-tune; a module must sit upstream \\
\addlinespace[2pt]
DeepScientist \citep{weng2025deepscientistadvancingfrontierpushingscientific} & Valuation against own findings memory (paper); prompt-file check (repo) & Hypothesis record; repo: draft template (claim, closest prior work, novelty type) & Own findings archive; literature via mandated sweep ($\ge$5 closest papers) & $\langle$utility, quality, exploration$\rangle$ 0--100 drives compute (paper); 0/1/2 novelty rubric, $<7$ blocks promotion (repo) & Yes (repo): editable markdown check; paper's scoring code not public \\
\addlinespace[2pt]
Robin \citep{ghareeb2025robinmultiagentautomatingscientific} & Novelty is one of five criteria in pairwise ranking, after per-candidate lit reports & Hypothesis name plus 800--1500w literature report; judged in pairs & Literature baked into the report upstream; parametric at judge time & Pairwise winner $\to$ Bradley--Terry strength; nothing discarded; orders for human wet-lab pick & Indirect: pool fixed before the judge; swap the pair comparison \\
\addlinespace[2pt]
AI-Researcher \citep{tang2025airesearcherautonomousscientificinnovation} & ``Pick the most novel of 5''; no similarity computed & Five markdown ideas (1--3k words) in one prompt & Seeded reference papers and repos on disk; search tools disabled & Free text: one selected idea report; the other four silently dropped; no score persisted & Yes: per-task reference list; replace the selection call \\
\addlinespace[2pt]
ARC \citep{liu2026autoresearchclawselfreinforcingautonomousresearch} & Stage-08 lexical check, explicitly non-blocking & Topic plus \texttt{hypotheses.md} (H1--H4, 300--1200w) plus seen-paper rows & Live search ($\le$5 queries $\times$ 15; OpenAlex, S2, arXiv) $\cup$ Stage-04 candidates; titles+abstracts & Keyword-Jaccard score 0--1, band, recommendation in \{proceed, differentiate, abort\}; advisory to a human & Yes: pass \texttt{papers\_already\_seen} or shadow \texttt{check\_novelty()} \\
\addlinespace[2pt]
SciMON \citep{wang-etal-2024-scimon} & Retrieve $\to$ compare $\to$ rewrite loop after initial generation & Idea plus background and problem context & Top-20 idea sentences from the training set (SentenceBERT) & LLM rewrites the idea until similarity falls below a threshold & Yes: all modules explicit \\
\addlinespace[2pt]
Idea Novelty Checker \citep{shahid-etal-2025-literature} & Generate $\to$ check $\to$ replace facet $\to$ regenerate & Idea decomposed into domain, purpose, mechanism, evaluation facets & Top-100 embedding-filtered papers (ideation, semantic search, related work), RankGPT top-10 & Binary label plus explanation and alternative facet; human may override & Yes: separate stages; new papers injectable before reranking \\
\addlinespace[2pt]
VirSci \citep{su2025headsbetteroneimproved} & Relative-novelty voting plus self-review of the finished abstract & Idea description, experiment, self-rating & Pre-cutoff papers (title, abstract, citations); per-idea retrieval & Preferred candidate plus reasoning; plurality vote selects the one idea written up & Yes: pool replaceable \\
\addlinespace[2pt]
AI Co-Scientist \citep{gottweis2026accelerating} & Fast initial review, then literature-grounded full review & Hypothesis or proposal plus parsed goal, constraints, and whether novelty is required & Parametric (initial); web search or scientist-provided repository (full) & Non-novel hypotheses discarded at initial review; full review summarizes known work as the comparison & Partly: provided pool replaceable, web search is not fixed \\
\bottomrule
\end{tabular}
\caption{Novelty judgments embedded in surveyed AI-scientist pipelines. Pool controllable: where an external module can replace the native judgment.}
\label{tab:survey-embedded}
\end{table*}

\begin{table*}[!htbp]
\centering
\scriptsize
\setlength{\tabcolsep}{3pt}
\begin{tabular}{p{2.2cm}p{1.9cm}p{3.5cm}p{3.6cm}p{3.5cm}}
\toprule
\rowcolor{gray!15}\textbf{System} & \textbf{Judges what} & \textbf{Input $I$} & \textbf{Pool} & \textbf{Output $O$} \\
\midrule
RAG-Novelty \citep{lin-etal-2025-evaluating} & Published papers & Title and abstract plus two retrieval scalars (mean neighbor date, mean cosine); neighbors never reach the prompt & Own vector store ($\approx$12k per field), $k{=}10$, pre-date filter; fully swappable & Integer 1--10 plus justification \\
\addlinespace[2pt]
OpenNovelty \citep{zhang2026opennovelty} & Full papers & PDF $\to$ core task and 1--3 contribution claims & Proprietary search API; top-50 per task, top-10 per contribution; citation index is the injection point & Per contribution: can refute / cannot refute / unclear, with evidence and summary; no numeric score \\
\addlinespace[2pt]
NovaScore \citep{ai-etal-2025-novascore} & News documents & Document $\to$ atomic content units & Unit bank (SBERT, top-5, $\ge$0.6); experimenter-defined & Per-unit binary novelty $\times$ salience $\to$ $[0,1]$ aggregate \\
\addlinespace[2pt]
InnoEval \citep{qiao2026innoevalresearchideaevaluation} & Research ideas & Six-tuple (TLDR, motivations, questions, methods, settings, results), auto-extracted & Six-source search, top-10 per category; swappable searchers & 0--10 per dimension incl.\ novelty, five personas, mean plus accept-tier label \\
\addlinespace[2pt]
ScholarEval \citep{moussa2025scholareval} & Research ideas & Research plan, with comparison dimensions extracted from it & Retrieved literature; swappable & Contribution arm: pairwise score against each retrieved paper per dimension, averaged to $-10$ to $+10$; no accept threshold \\
\bottomrule
\end{tabular}
\caption{Standalone novelty judges.}
\label{tab:survey-standalone}

\vspace{1.2em}
\setlength{\tabcolsep}{3pt}
\begin{tabular}{p{2.1cm}p{6.2cm}p{3.6cm}p{3.4cm}}
\toprule
\rowcolor{gray!15}\textbf{System} & \textbf{Committed idea artifact (post-check, pre-execution)} & \textbf{Where captured} & \textbf{What runs on it} \\
\midrule
AI Scientist v1 & Idea dict (Name, Title, Experiment), novelty-filtered & \texttt{ideas.json} per template & experiments $\to$ full paper \\
AI Scientist v2 & Idea JSON incl.\ Short Hypothesis, Related Work, Abstract, Experiments & ideas JSON; human picks an index & tree-search experiments $\to$ paper \\
CoI-Agent & \texttt{final\_idea} free-text blob & return value of the regeneration loop & experiment design \\
Si et al. & Full proposal JSON (7 fields incl.\ test cases, fallback plan) & \texttt{top\_ideas.json}, rank-ordered and filter-passed & human study / execution study \\
ResearchAgent & Problem, method, experiment design, each with rationale; best-of-$N$ & pipeline history, max-score pick & (idea is the end product) \\
Agent Laboratory & Research plan from plan formulation & agent state & experiments $\to$ paper \\
SCI-IDEA & Refined ``Aha'' idea text & module-2 output & (idea is the end product) \\
CycleResearcher & None exposed; output is a full paper & LaTeX & review $\to$ RL reward \\
DeepScientist & Promoted hypothesis record (claim plus valuation vector) & findings memory & expensive validation runs \\
Robin & Ranked candidates (name, lit report, strength score) & ranked CSV & human wet-lab selection \\
AI-Researcher & One selected idea report (1--3k words) & selection-call output & code survey $\to$ implementation \\
ARC & \texttt{hypotheses.md} (H1--H4 with rationale, prediction, failure condition) & stage-08 artifact & later stages $\to$ report \\
\bottomrule
\end{tabular}
\caption{What each pipeline passes downstream after its novelty check.}
\label{tab:survey-committed}
\end{table*}

\begin{table}[!htbp]
\centering
\scriptsize
\setlength{\tabcolsep}{3pt}
\begin{tabular}{>{\raggedright\arraybackslash}p{2.5cm}>{\raggedright\arraybackslash}p{4.7cm}}
\toprule
\rowcolor{gray!15}\textbf{System} & \textbf{Reason} \\
\midrule
\multicolumn{2}{l}{\emph{Included}} \\
AI Scientist v1, CoI-Agent, Si et al., ARC & deployed check \\
\addlinespace[2pt]
ResearchAgent, AI-Scientist-v2 & judgment extracted from the pipeline \\
\addlinespace[2pt]
ScholarEval & standalone, but scores an idea against a supplied paper set \\
\midrule
\multicolumn{2}{l}{\emph{Excluded}} \\
SCI-IDEA & no code released \\
\addlinespace[2pt]
CycleResearcher & monolithic fine-tune, no separable judgment \\
\addlinespace[2pt]
DeepScientist & paper-era judge prompt unpublished \\
\addlinespace[2pt]
Agent Laboratory & no check; its reviewer scores the finished paper \\
\addlinespace[2pt]
Robin, AI-Researcher, VirSci & relative selection among candidates, no per-idea verdict \\
\addlinespace[2pt]
AI Co-Scientist & closed source, web search not fixable \\
\addlinespace[2pt]
SciMON & pool is training-set idea sentences; output is a rewrite \\
\addlinespace[2pt]
Other standalone judges & expect a published paper or PDF; non-binary outputs \\
\bottomrule
\end{tabular}
\caption{Systems included in and excluded from the audit.}
\label{tab:survey-select}
\end{table}

\begin{table}[!htbp]
\centering
\scriptsize
\setlength{\tabcolsep}{3pt}
\begin{tabular}{p{2.6cm}p{4.6cm}}
\toprule
\rowcolor{gray!15}\textbf{Check} & \textbf{Native input built from (title, description)} \\
\midrule
AI Scientist v1 & idea dict: Name = slugified title, Title = title, Experiment = description \\
CoI-Agent & one idea blob, ``title: description'' \\
Si et al. & one idea blob, ``title: description''; title also used as a search query \\
ARC & title and description together as the hypothesis text \\
ResearchAgent & Problem = title, Rationale = description; target paper = title and description \\
ScholarEval & research plan = ``title: description'' \\
AI-Scientist-v2 & Title = title, Proposal = description, in the reflection prompt \\
\bottomrule
\end{tabular}
\caption{How the shared proposal is passed to each check's native input.}
\label{tab:survey-adapters}
\end{table}

\begin{table}[!htbp]
\centering
\footnotesize
\setlength{\tabcolsep}{4pt}
\begin{tabular}{lrr}
\toprule
\rowcolor{gray!15}\textbf{Call} & \textbf{\$ / proposal} & \textbf{Latency} \\
\midrule
\textsc{ClaimNov} verdict & 0.010 & 2.6\,s \\
CoI-Agent check & 0.039 & 19\,s \\
ARC (lexical) & 0 & --- \\
\bottomrule
\end{tabular}
\caption{Cost per proposal at list price and median latency of the final judge call (Sonnet~4.6), measured over 948 \textsc{ClaimNov} and 446 CoI-Agent calls. ARC makes no model call.}
\label{tab:cost}
\end{table}

\FloatBarrier
\section{Adapter Details}
\label{app:adapters}
We pass the same (proposal, pool) pair to each check. For all checks that do some kind of search, we sandbox the search, ranking the pool by cosine similarity (\embmodel{}) to the query (checks can still use their native query style and specified depth $k$). In all cases, we set the final judge to Sonnet~4.6 at temperature 0 (Table~\ref{tab:adapter-rules}). In the OpenNovelty-style baseline, we reimplement the per-contribution design of \citet{zhang2026opennovelty} with BM25 retrieval to avoid sharing a retriever with \textsc{ClaimNov}. BM25 throughout is \texttt{rank\_bm25}'s BM25Okapi with default parameters ($k_1{=}1.5$, $b{=}0.75$) over lowercased alphanumeric tokens with stopwords removed, and the scientific embeddings are SciNCL (\texttt{malteos/scincl}) and SPECTER2 (\texttt{allenai/specter2\_base} with the proximity adapter).

\begin{table}[!htbp]
\centering
\scriptsize
\setlength{\tabcolsep}{3pt}
\begin{tabular}{>{\raggedright\arraybackslash}p{1.5cm}>{\raggedright\arraybackslash}p{2.4cm}>{\raggedright\arraybackslash}p{3.2cm}}
\toprule
\rowcolor{gray!15}\textbf{Check} & \textbf{Search into pool} & \textbf{Decision rule} \\
\midrule
AI Sci.\ v1 & up to 10 rounds, each one LLM query $\to$ top-10 & ``Decision made: not novel'' rejects; no decision after 10 rounds rejects (shipped default) \\
CoI-Agent & up to 3 LLM queries $\to$ top-5 each, deduplicated & \texttt{<similar>1} rejects and names a paper; empty retrieval passes (shipped) \\
Si et al. & 1--3 keyword queries plus the title $\to$ top-10 each & papers scored 1--10, the top 5 checked one by one, any ``same problem and approach'' rejects \\
ARC & whole pool, no LLM & keyword Jaccard and title similarity $\ge 0.25$ for some paper rejects \\
ResearchAgent & proposal $\to$ top-10 titles & Originality rating $\le 3$ of 5 rejects (no native threshold; shipped temperature 1.25 not kept) \\
ScholarEval & proposal $\to$ top-8 & mean pairwise score over the 8 papers $\le 0$ rejects (no native threshold) \\
AI-Sci.-v2 & proposal $\to$ top-10 & reflection prompt plus an appended \texttt{NOVELTY: pass/fail} slot \\
OpenNov.-style & 1--3 extracted contributions $\to$ BM25 top-10 each & one judge call over all contributions; rejects when every contribution is \texttt{can\_refute} (rejecting on any raises false alarms to 55.8\%) \\
\bottomrule
\end{tabular}
\caption{Sandboxed search and decision rule of each adapter.}
\label{tab:adapter-rules}
\end{table}

\FloatBarrier
\section{\textsc{ClaimNov} Prompts}
\label{app:prompts}
The exact \textsc{ClaimNov} prompts.

\begin{tcolorbox}[modprompt, title=Claim Extraction (system prompt)]
\ttfamily\small
\setlength{\parskip}{4pt}\setlength{\parindent}{0pt}
\raggedright
You extract the contribution/novelty claims from the abstract of a scientific paper. A 'claim' is one atomic, headline contribution the authors make — what they would put in a numbered 'Our contributions:' list.

BE CONSERVATIVE. Most papers have 1-4 atomic contributions, not 5-8. Prefer one consolidated claim over fine-grained sub-claims.

Dimension definitions:\\
- methodology: a new algorithm, architecture, training procedure, loss function, selection mechanism, or technique that is the paper's primary *solution* to a problem. The contribution IS the method — how something is done. Includes objective functions, encoders, and model claims when they are the central proposed technique.\\
- result: a new empirical or theoretical finding, analysis insight, performance observation, or theoretical characterization that is discovered/proven rather than proposed as a solution. This includes: showing that an existing design choice is effective, demonstrating generalization ability, reporting accuracy/fairness/efficiency gains, and deriving theoretical distributions or bounds when the distribution/bound is the finding, not a technique being proposed.\\
- resource: a new dataset, benchmark, tool, model release, or codebase released by THIS paper.\\
- problem: a new problem formulation, task, setting, scenario, or research question this paper introduces — the question or framing, not the answer. Includes formalizations of novel learning settings, new task definitions, and new evaluation scenarios.

Dimension boundary rules (apply these when confused):\\
- An OBJECTIVE FUNCTION or LOSS is `methodology` when it is the paper's proposed technique for solving a problem; it is `result` only when it is a derived theoretical characterization that the paper proves as a finding.\\
- A DISTRIBUTION (e.g., Laplace on SO(3)) that is the paper's main proposed modeling tool is `methodology`; if the paper's claim is \textquotedbl{}we prove X has property Y\textquotedbl{} with no proposed use, it may be `result`.\\
- \textquotedbl{}We show / demonstrate / find that technique T is effective\textquotedbl{} → `result`, not `methodology`. Effectiveness findings about a claim are results even if the claim is also described elsewhere.\\
- A FRAMEWORK or FORMALIZATION that characterizes existing assumptions, explains prior work, or provides theoretical grounding (rather than solving a new problem) is `methodology` when it is a new analytical tool, NOT `problem`.\\
- A DATASET or BENCHMARK constructed as part of the paper → `resource`, not `methodology`, even if construction required significant effort.\\
- A SETTING, TASK, or SCENARIO introduced for the first time → `problem`, even when phrased in proposal voice (\textquotedbl{}we propose the setting of…\textquotedbl{}).\\
- When a paper both introduces a new setting/task AND a new method for it, extract BOTH as separate claims (`problem` + `methodology`).

Decomposition rules:\\
- DO split contributions that are independently verifiable AND of different dimensions.\\
- DO NOT split a single method into sub-claims, or a single result into sub-numbers.\\
- DO NOT extract motivation/background or claims about prior work.\\
- DO extract a `problem` claim whenever the paper introduces a novel task, setting, or formalization — even if it also proposes a method for it. These are separate contributions of different dimensions.\\
- DO extract a `result` claim whenever the abstract reports empirical outcomes, performance gains, or evaluative findings — even if the paper's main contribution is a method.\\
- DO extract a `resource` claim whenever the paper releases a dataset, benchmark, or tool — even as a secondary contribution.

Recall guidance — scan for these contribution signals before finalising:\\
- Is there a novel problem/task/setting/formalization introduced? → `problem`\\
- Is there a new algorithm, architecture, training procedure, or loss function? → `methodology`\\
- Are benchmarks, datasets, or tools released? → `resource`\\
- Are empirical or theoretical results reported (accuracy gains, effectiveness demonstrations, theoretical bounds, generalization experiments)? → `result`\\
- Does the abstract describe a claim or design choice shown to be effective (e.g., encoder choice, selection mechanism, initialization)? → likely a `result` claim unless it is the central proposed technique.\\
- Does the abstract describe a method with multiple named sub-claims (e.g., a selection mechanism, a refinement stage)? → consider whether each is an independently verifiable methodological contribution or whether they together form one consolidated `methodology` claim.\\
- Does the paper introduce a novel scenario or learning setting that prior work has not formalized? → `problem` (check even when the abstract's primary voice is about a proposed method).

Proposal-voice caveats (important):\\
- \textquotedbl{}We propose/introduce X\textquotedbl{} where X is a SETTING, TASK, SCENARIO, or problem formulation is dimension `problem`, NOT methodology — even when phrased in proposal voice. `methodology` is reserved for the technique/algorithm that solves a problem.\\
- \textquotedbl{}We propose/introduce X\textquotedbl{} where X is a METHOD, ALGORITHM, OBJECTIVE, or TECHNIQUE is dimension `methodology`.\\
- Future-tense proposal wording (\textquotedbl{}we will develop\textquotedbl{}, \textquotedbl{}we plan to\textquotedbl{}) does not change the dimension: classify by what the contribution IS, not how it is voiced.

For each claim also extract its facet:\\
- attribute: the comparable axis of the claim, as a short noun phrase (the retrieval key).\\
- value: what the paper asserts on that axis.

Return JSON:\\
\{\textquotedbl{}claims\textquotedbl{}: [\{\textquotedbl{}claim\textquotedbl{}: \textquotedbl{}\textless{}one self-contained sentence\textgreater{}\textquotedbl{}, \textquotedbl{}dimension\textquotedbl{}: \textquotedbl{}methodology|result|resource|problem\textquotedbl{}, \textquotedbl{}attribute\textquotedbl{}: \textquotedbl{}...\textquotedbl{}, \textquotedbl{}value\textquotedbl{}: \textquotedbl{}...\textquotedbl{}, \textquotedbl{}abstract\_span\textquotedbl{}: \textquotedbl{}\textless{}verbatim quote\textgreater{}\textquotedbl{}\}]\}

If the abstract contains no contribution claims, return \{\textquotedbl{}claims\textquotedbl{}: []\}.
\end{tcolorbox}

\begin{tcolorbox}[modprompt, title=Support Counting (Stage B)]
\ttfamily\small
\setlength{\parskip}{4pt}\setlength{\parindent}{0pt}
\raggedright
You find ALL the supporting evidence in a scientific paper's body for a specific contribution claim from its abstract. Select EVERY paragraph (by index) containing direct supporting evidence — descriptions, numbers, theorem statements, implementation details, appendix analysis tied to the claim. There is NO cap. Skip only paragraphs not actually about this claim (background, related work, other contributions).

Order the selected paragraphs by relevance to the claim, most relevant first.

Return JSON: \{\textquotedbl{}evidence\textquotedbl{}: [\{\textquotedbl{}paragraph\_index\textquotedbl{}: \textless{}int\textgreater{}\}]\}
\end{tcolorbox}

\begin{tcolorbox}[modprompt, title=Graded Entailment Judge]
\ttfamily\small
\setlength{\parskip}{4pt}\setlength{\parindent}{0pt}
\raggedright
You are evaluating claim entailment for research-paper novelty analysis.

Your task is to decide the level to which Claim 2 entails or covers Claim 1.\\
The context blocks are only additional context. They are NOT the claims being evaluated directly.\\
Evaluate entailment between Claim 2 and Claim 1.\\
Use the 1-5 scale below.

Entailment levels:\\
1 = No entailment / unrelated. Claim 2 does not cover Claim 1.\\
2 = Weak overlap. Claim 2 shares a topic with Claim 1 but does not really support the same contribution.\\
3 = Partial entailment. Claim 2 covers part of Claim 1, but important method/result/resource details differ or are missing.\\
4 = Strong entailment. Claim 2 mostly covers Claim 1, with only minor missing details or wording differences.\\
5 = Full entailment. Claim 2 fully covers the substantive content of Claim 1.

When the claims assert numeric values: a prior claim anticipates a target only if it establishes the same kind of finding under comparable conditions. For performance/achievement claims on a shared benchmark and metric, a prior result at least as strong may fully entail the target. For measurements, effect sizes, estimates, or findings in different settings or populations, a different (even larger) prior number does NOT by itself entail the target — the direction, setting, or refined estimate may be exactly what is new.

Claim 1 (target):\\
\textless{}target claim\textgreater{}

Additional context for Claim 1:\\
Facet — attribute: \textless{}attribute\textgreater{}; value: \textless{}value\textgreater{}

Claim 2 (prior work: \textless{}prior title\textgreater{}):\\
\textless{}prior claim\textgreater{}

Additional context for Claim 2:\\
Facet — attribute: \textless{}attribute\textgreater{}; value: \textless{}value\textgreater{}

Output ONLY one integer: 1, 2, 3, 4, or 5.
\end{tcolorbox}

\begin{tcolorbox}[modprompt, title=Verdict (profile judge)]
\ttfamily\small
\setlength{\parskip}{4pt}\setlength{\parindent}{0pt}
\raggedright
You are the final novelty judge for a proposed research IDEA. A claim-level analysis has already been run: the idea's contribution claims were extracted, and for each claim the most similar existing claims were retrieved from a pool of prior papers and rated for how far they cover the idea's claim (1 = unrelated, 2 = weak topical overlap, 3 = partial coverage with important details differing, 4 = strong coverage with minor differences, 5 = full coverage of the claim's substantive content).

Decide whether the IDEA is already done: is there a paper in the profile that does substantially what the idea proposes, same core problem AND same core approach? Judge the idea as a whole: an anticipated peripheral claim does not make the idea non-novel, but an anticipated core contribution does, even if side claims are new. Superficial topic overlap or related prior work that the idea builds on = novel.

IDEA:\\
\{idea\}

NOVELTY PROFILE:\\
\{for each claim: the claim and its dimension; under it, up to five matched prior claims as [paper title] claim text -> coverage k/5\}

PAPERS IN THE PROFILE:\\
\{title and abstract of each paper appearing above, up to eight\}

Return JSON only:\\
\{\textquotedbl{}decision\textquotedbl{}: \textquotedbl{}novel\textquotedbl{} | \textquotedbl{}not\_novel\textquotedbl{}, \textquotedbl{}anticipating\_paper\textquotedbl{}: \textquotedbl{}<exact paper title or null>\textquotedbl{}, \textquotedbl{}established\_claims\textquotedbl{}: [<claim indices the paper already establishes>], \textquotedbl{}why\textquotedbl{}: \textquotedbl{}<one or two sentences>\textquotedbl{}\}
\end{tcolorbox}

\FloatBarrier
\section{Disguise Prompts}
\label{app:disguise-prompts}

We provide the exact prompts used to disguise proposals for each rung, as well as the prompt used for the verifier for rungs 3 and 4. Note the rung-1 prompt is also used to rewrite claim-disputed papers (\S\ref{sec:materials}). These are run with Sonnet~4.6.

\begin{tcolorbox}[modprompt, title={Rung 1: voice swap}]
\ttfamily\small
\setlength{\parskip}{4pt}\setlength{\parindent}{0pt}
\raggedright
You are drafting a research idea proposal. Below is the abstract of a completed research paper. Recast it as a NEW research idea someone is proposing to work on: write a proposal title and a $\sim$200-word description in proposal voice (future tense, \textquotedbl{}we propose / we will\textquotedbl{}). Keep the technical content as close to the original as possible — same contribution, same method, same expected findings; only the framing changes from \textquotedbl{}we did\textquotedbl{} to \textquotedbl{}we propose\textquotedbl{}.

Abstract:\\
\{abstract\}

Return JSON only: \{\textquotedbl{}title\textquotedbl{}: \textquotedbl{}...\textquotedbl{}, \textquotedbl{}description\textquotedbl{}: \textquotedbl{}...\textquotedbl{}\}
\end{tcolorbox}

\begin{tcolorbox}[modprompt, title={Rung 2: deep paraphrase}]
\ttfamily\small
\setlength{\parskip}{4pt}\setlength{\parindent}{0pt}
\raggedright
You are drafting a research idea proposal. Below is the abstract of a completed research paper. Recast it as a NEW research idea someone is proposing to work on, as a deep paraphrase: preserve every technical claim, the method, and the expected findings, but rewrite from scratch in entirely different vocabulary and sentence structure — no phrase over 4 words may match the original. Technical proper nouns and dataset names may remain. Write a proposal title (different wording from the original title) and a $\sim$200-word description in proposal voice.

Abstract:\\
\{abstract\}

Return JSON only: \{\textquotedbl{}title\textquotedbl{}: \textquotedbl{}...\textquotedbl{}, \textquotedbl{}description\textquotedbl{}: \textquotedbl{}...\textquotedbl{}\}
\end{tcolorbox}

\begin{tcolorbox}[modprompt, title={Rung 3: reframe}]
\ttfamily\small
\setlength{\parskip}{4pt}\setlength{\parindent}{0pt}
\raggedright
You are drafting a research idea proposal. Below is the abstract of a completed research paper. Recast it as a NEW research idea proposal that makes the SAME core technical contribution but is presented completely differently:\\
- motivate it from a different angle (different application domain or research concern than the original opens with)\\
- reorder the contributions; do not mirror the original's sentence structure or presentation order\\
- do not reuse the original's method name; either leave the method unnamed or give it a neutral descriptive name\\
- do not reuse the original's examples or illustrative scenarios\\
- keep every substantive technical claim: the same method, the same key mechanism, the same expected findings (numbers may be stated approximately)\\
$\sim$200 words, proposal voice (future tense). Title must not echo the original title.

Abstract:\\
\{abstract\}

Return JSON only: \{\textquotedbl{}title\textquotedbl{}: \textquotedbl{}...\textquotedbl{}, \textquotedbl{}description\textquotedbl{}: \textquotedbl{}...\textquotedbl{}\}
\end{tcolorbox}

\begin{tcolorbox}[modprompt, title={Rung 4: adversarial restructuring}]
\ttfamily\small
\setlength{\parskip}{4pt}\setlength{\parindent}{0pt}
\raggedright
You are drafting a research idea proposal. Below is the abstract of a completed research paper. Write a proposal that a hurried reader would take for an unrelated new idea, but that an expert doing a careful side-by-side comparison would judge to be substantially the same work:\\
- start from a different theoretical or practical starting point and let the core mechanism \textquotedbl{}emerge\textquotedbl{} as the proposed approach\\
- rename or leave unnamed every method/system; drop or generalize all specific numbers (\textquotedbl{}substantial improvement\textquotedbl{}, not percentages)\\
- restructure completely: different opening problem, different order, different emphasis (e.g., foreground what the original treats as secondary)\\
- you may add plausible framing (motivation, evaluation plans) but must NOT add or change any substantive technical contribution\\
- the same core problem must be solved by the same core mechanism with the same expected outcome\\
$\sim$200 words, proposal voice. Title must appear unrelated to the original.

Abstract:\\
\{abstract\}

Return JSON only: \{\textquotedbl{}title\textquotedbl{}: \textquotedbl{}...\textquotedbl{}, \textquotedbl{}description\textquotedbl{}: \textquotedbl{}...\textquotedbl{}\}
\end{tcolorbox}

\begin{tcolorbox}[modprompt, title={Verifier (rungs 3 and 4)}]
\ttfamily\small
\setlength{\parskip}{4pt}\setlength{\parindent}{0pt}
\raggedright
Do these two texts describe substantially the same core research contribution — same core problem solved by the same core mechanism? Answer \textquotedbl{}yes\textquotedbl{} or \textquotedbl{}no\textquotedbl{}, then one sentence.

TEXT A (paper abstract):\\
\{abstract\}

TEXT B (proposal):\\
\{description\}
\end{tcolorbox}

\FloatBarrier
\section{Later-Realized Labeling Prompts}
\label{app:label-prompts}
We provide exact prompts used to label the later-realized set, run with Opus~4.6.

\begin{tcolorbox}[modprompt, title=Contribution List (independent extraction)]
\ttfamily\small
\setlength{\parskip}{4pt}\setlength{\parindent}{0pt}
\raggedright
Write the \textquotedbl{}Our contributions are:\textquotedbl{} list that this proposal's authors would put in their introduction, one contribution per line. Output nothing else.

PROPOSAL:\\
\{title: description\}
\end{tcolorbox}

\begin{tcolorbox}[modprompt, title={Per-Contribution Label (run once against the pool, once against $F$)}]
\ttfamily\small
\setlength{\parskip}{4pt}\setlength{\parindent}{0pt}
\raggedright
You are labeling a proposed research IDEA at the level of its individual contribution claims.

IDEA:\\
\{title: description\}

IDEA'S CONTRIBUTION CLAIMS:\\
\{the numbered contribution list from the prompt above\}

CANDIDATE PAPERS (\{published before the cutoff | published after the cutoff\}):\\
\{title and abstract of the 8 nearest papers in the pool, or in $F$\}

For EACH claim, decide whether some candidate paper already establishes it: the same thing is done or shown for that claim (same core problem and same core approach for that claim), not merely related work or work the claim builds on. If yes, name the paper.

Return JSON only:\\
\{\textquotedbl{}claims\textquotedbl{}: [\{\textquotedbl{}i\textquotedbl{}: <claim index>, \textquotedbl{}established\textquotedbl{}: \textquotedbl{}yes\textquotedbl{}|\textquotedbl{}no\textquotedbl{}, \textquotedbl{}by\textquotedbl{}: \textquotedbl{}<exact paper title or null>\textquotedbl{}, \textquotedbl{}why\textquotedbl{}: \textquotedbl{}<one sentence>\textquotedbl{}\}]\}
\end{tcolorbox}

\FloatBarrier
\section{Label Validation}
\label{app:labels}
For $n{=}100$ pairs of papers $A$ and $B$ where an LLM identified a prior-art dispute, two annotators independently mark (i) whether the dispute is genuine, (ii) whether the extracted $B$ is the correct prior art, and (iii) the dispute type. They achieve (i) 81/100 raw agreement with Cohen's $\kappa = 0.62$, (ii) 73/100 and $\kappa{=}0.47$, and (iii) 52/100 and $\kappa{=}0.40$, and we drop the dispute-type labels due to the low kappa value. A third annotator then marked the 63 rows with any disagreements, blinded to the labels of the first two annotators. By 2-of-3 majority we find that we have 59/100 genuine disputes with 67/100 having correctly-identified $B$. We use the 40 pairs that have a genuine dispute and correct $B$ as the label-clean subset, 36 of which have usable pools (Table~\ref{tab:label-clean}). All three annotators have graduate-level education in English in the relevant research field.

\begin{table}[!htbp]
\centering
\footnotesize
\begin{tabular}{lcc}
\toprule
\rowcolor{gray!15}\textbf{Question} & \textbf{Agreement} & $\bm{\kappa}$ \\
\midrule
Dispute is genuine & 81/100 & .62 \\
$B$ is the correct prior art & 73/100 & .47 \\
Dispute type (dropped) & 52/100 & .40 \\
\bottomrule
\end{tabular}
\caption{Raw agreement and Cohen's $\kappa$ between the first two annotators.}
\label{tab:label-agree}
\end{table}

\begin{table}[!htbp]
\centering
\footnotesize
\setlength{\tabcolsep}{4pt}
\begin{tabular}{lccc}
\toprule
\rowcolor{gray!15}\textbf{Check} & \textbf{Rej.\ acc.} & \textbf{Rej.\ rej.} & \textbf{Names $B$} \\
\midrule
AI Scientist v1 & 7/16 & 7/19 & --- \\
CoI-Agent & 2/16 & 3/19 & --- \\
Si et al. & 14/16 & 14/19 & 19, rejecting \\
OpenNov.-style & 2/16 & 3/19 & --- \\
\midrule
\textsc{ClaimNov} & 1/16 & 1/19 & 15, 14 passing \\
\bottomrule
\end{tabular}
\caption{Rejections on the 36 label-clean pairs. Names $B$: for \textsc{ClaimNov}, $B$ in the profile at level 3 or above.}
\label{tab:label-clean}
\end{table}

We give the annotation instructions below. The audit of later-realized and sorter-matched ideas uses the second set.

\begin{tcolorbox}[modprompt, title={Claim-dispute annotation (annotators 1 and 2)}]
\small
\setlength{\parskip}{4pt}\setlength{\parindent}{0pt}
\raggedright
Fill in your file completely and independently, with no discussion of any row until both files are done. Judge only from what is in the row (plus the linked PDFs). Do not search the literature yourself. Mark any row you are unsure about as low confidence and give the reason in the notes.

For each row, read A's title and abstract, the reviewer's snippet, and B's title and abstract, then answer the following.

\textbf{Q1. Is the reviewer genuinely disputing A's novelty by pointing at prior work?}
\emph{Yes} if the reviewer asserts or clearly implies that A's idea, method or setting already exists in prior work (``this was already done'', ``very similar to [B]'', ``unclear how this differs from [B]'', ``the claim of being first is not true''). \emph{No} for a neutral citation, a request to add a citation, background discussion, praise, or a comparison request that does not question novelty.

\textbf{Q2. Is the listed B the work the reviewer is referring to?}
\emph{Yes} if B's title and identity match the work invoked in the snippet. \emph{No} if the wrong paper was linked (e.g., same author, different paper). \emph{Cannot assess} if B is unresolvable shorthand that the snippet does not identify.

\textbf{Q3. What kind of dispute is it?} (Answer even if Q1 is no, using your best reading.)
\emph{duplicate}: essentially the same thing was already done. \emph{substantial overlap}: the core contribution heavily overlaps B, though not identically. \emph{incremental}: A is a minor delta over B. \emph{framing}: the reviewer contests a novelty statement or overclaim rather than the contribution itself. \emph{other}: explain in the notes.

A reviewer who concedes the contribution while noting that the problem area is old is \emph{framing}, not \emph{duplicate}. A dispute about the problem setting being identical is \emph{substantial overlap}.
\end{tcolorbox}

\begin{tcolorbox}[modprompt, title={Claim-dispute annotation (annotator 3, blind)}]
\small
\setlength{\parskip}{4pt}\setlength{\parindent}{0pt}
\raggedright
Same task and instructions as above, for Q1 and Q2. You are the third independent annotator. These are the pairs where the first two annotators disagreed, but their answers are not shown and the row order is shuffled. The majority vote is computed after you return the file.
\end{tcolorbox}

\begin{tcolorbox}[modprompt, title={Later-realized audit}]
\small
\setlength{\parskip}{4pt}\setlength{\parindent}{0pt}
\raggedright
Each idea comes with the 20 most similar papers from the prior-work pool, all published before the cutoff, most similar first (P1--P20). Some items also list one or more later papers published after the cutoff, marked ``F''. These are for reference only. Do not count an F paper as prior work. Work independently, do not search the web, and judge only from the idea text and the listed abstracts.

\textbf{Q1. Does any listed paper (P1--P20) already do substantially what the idea proposes, with the same core problem and the same core approach?}
Answer yes or no, and if yes, give the number of the paper that does it best. Answer no for a paper that only shares the topic, that the idea builds on, or that does one side component of the idea. Answer yes only if the idea's main contribution is already there. A missing peripheral detail does not make the idea new.

\textbf{Q2 (only if Q1 is no). Is any single contribution of the idea clearly established by a listed paper, even though the idea as a whole is not?}
Answer yes or no, and if yes, give the paper number. A few words naming the contribution are optional.

\textbf{Confidence.} Rate your Q1 answer 1 (guess), 2 (fairly sure) or 3 (certain).
\end{tcolorbox}

\subsection{Audit of generated ideas}
\label{app:audit}
We audit 60 generated ideas by hand with the second set of instructions above, one annotator per item. These are 45 later-realized ideas (25 at random, 10 that the whole-idea Opus pass called already done, and 10 that a low-false-alarm check rejected) and 15 sorter-matched ideas. This labeling is extremely strenuous, since each item requires reading an idea against 20 abstracts and deciding whether any one paper does substantially the same thing, a judgment on which experts often disagree \citep{si2025llms}. Annotators confirm 28 of the 45 later-realized ideas as novel relative to the pool (a stratum-weighted 60\% of all 400) and judge 9 of the 15 sorter-matched ideas already done. Table~\ref{tab:audit} gives each check's rejections on both groups of later-realized ideas. On the 9 sorter-matched ideas judged already done, \textsc{ClaimNov} with fixed retrieval depth rejects 1 and the w/o-profile judge 7, which also rejects 3 of the 6 judged novel.

\begin{table}[!htbp]
\centering
\footnotesize
\setlength{\tabcolsep}{4pt}
\begin{tabular}{lrr}
\toprule
\rowcolor{gray!15}\textbf{Check} & \textbf{Novel (28)}$\downarrow$ & \textbf{Done (17)} \\
\midrule
AI Scientist v1 & 11 & 11 \\
CoI-Agent & 0 & 2 \\
Si et al. & 24 & 17 \\
ARC & 0 & 0 \\
ResearchAgent & 27 & 17 \\
ScholarEval & 0 & 0 \\
AI-Sci.-v2 & 14 & 10 \\
OpenNov.-style & 8 & 2 \\
W/o profile & 6 & 2 \\
\midrule
\textsc{ClaimNov} & 2 & 0 \\
\bottomrule
\end{tabular}
\caption{Rejections of the later-realized ideas annotators confirm as novel and of those they judge already done.}
\label{tab:audit}
\end{table}

\FloatBarrier
\section{Attribution by Dimension and Relation Type}
\label{app:attribution}
Table~\ref{tab:attr-dim} breaks down the claim-level attribution of \S\ref{sec:claim} by the disputed claim's dimension and by the relation type in the review.

\begin{table}[!htbp]
\centering
\footnotesize
\setlength{\tabcolsep}{3pt}
\begin{tabular}{lrrrr}
\toprule
\rowcolor{gray!15}\textbf{Disputed claim} & $\bm{n}$ & \textbf{extracted}$\uparrow$ & \textbf{$B$ in profile}$\uparrow$ & \textbf{lvl$\ge$3}$\uparrow$ \\
\midrule
Problem & 37 & 19 & 30 & 17 \\
Methodology & 42 & 42 & 29 & 17 \\
Resource & 39 & 12 & 20 & 6 \\
Result & 42 & 35 & 25 & 18 \\
\midrule
Prior art & 53 & --- & 37 & 23 \\
Comparison & 86 & --- & 49 & 24 \\
Overlap & 21 & --- & 18 & 11 \\
\bottomrule
\end{tabular}
\caption{Attribution on the claim-disputed set by dimension of the disputed claim (top) and by relation type (bottom). Extracted: a claim was extracted in that dimension.}
\label{tab:attr-dim}
\end{table}

\FloatBarrier
\section{\textsc{ClaimNov} Ablations}
\label{app:ablations}
We explain the ablations we evaluate for \textsc{ClaimNov}. Each aims to ablate one stage, leaving the rest unchanged (Table~\ref{tab:ablations}).

\begin{table}[!htbp]
\centering
\scriptsize
\setlength{\tabcolsep}{3pt}
\begin{tabular}{>{\raggedright\arraybackslash}p{1.9cm}>{\raggedright\arraybackslash}p{5.3cm}}
\toprule
\rowcolor{gray!15}\textbf{Ablation} & \textbf{Change from full \textsc{ClaimNov}} \\
\midrule
W/o claim profile & the verdict stage (same model, prompt) reads eight pool abstracts nearest to the proposal by embedding instead of the novelty profile \\
W/o verdict judge (graded check) & fail when aggregate score $N(P) < 0.5$ or any claim is covered at level 4 or higher by one prior claim \\
Per-claim verdict rule & one judge call per candidate paper on whether it establishes the central claim; any yes fails \\
Fixed depth ($k{=}10$) & the 10 nearest same-dimension prior claims for every claim, disregarding pool size \\
\bottomrule
\end{tabular}
\caption{\textsc{ClaimNov} ablations.}
\label{tab:ablations}
\end{table}

\FloatBarrier
\section{Disguises Caught by Rung}
\label{app:rungs}
Table~\ref{tab:rungs} breaks down Table~\ref{tab:catchfa} by disguise rung. Aside from ARC (which we expect to have such behavior since it's a fully lexical check), we don't observe significant fluctuation in catch for different checks across the rungs.

\begin{table}[!htbp]
\centering
\footnotesize
\begin{tabularx}{\linewidth}{lRRRR}
\toprule
\rowcolor{gray!15}\textbf{Check} & \textbf{r1}$\uparrow$ & \textbf{r2}$\uparrow$ & \textbf{r3}$\uparrow$ & \textbf{r4}$\uparrow$ \\
\midrule
AI Scientist v1 & \textbf{100} & \textbf{100} & \textbf{100} & \textbf{100} \\
CoI-Agent & \textbf{100} & \underline{99} & 96 & \underline{93} \\
Si et al. & \textbf{100} & \textbf{100} & \textbf{100} & \textbf{100} \\
ARC & 90 & 0 & 0 & 0 \\
ResearchAgent & 77 & 91 & 94 & 92 \\
ScholarEval & 5 & 5 & 5 & 5 \\
AI-Scientist-v2 & 94 & 96 & 97 & 92 \\
\midrule
\rowcolor{gray!25}\textbf{\textsc{ClaimNov}} & \underline{99} & 97 & \underline{98} & 92 \\
\bottomrule
\end{tabularx}
\caption{Disguises caught per rung (100 each, all 400 disguises). Bold: best, underline: second.}
\label{tab:rungs}
\end{table}

\FloatBarrier
\section{Intervals and Paired Tests}
\label{app:stats}
In Table~\ref{tab:ci} we give 95\% intervals for Table~\ref{tab:catchfa}. In Table~\ref{tab:mcnemar} we give exact McNemar tests. Finally, in Table~\ref{tab:stability} we give repeated runs and clustered tests. Main-table rows come from one run per check on the first-party Anthropic API. Later Claude runs (repeats, profile-plus-abstracts, anchor-in, larger pools) use Sonnet~4.6 on Amazon Bedrock. Tables~\ref{tab:ir} and~\ref{tab:levels} score attribution as retrieval.

\begin{table}[!htbp]
\centering
\footnotesize
\setlength{\tabcolsep}{2.5pt}
\resizebox{\columnwidth}{!}{\begin{tabular}{lccc}
\toprule
\rowcolor{gray!15}\textbf{Check} & \textbf{FA (\%)} & $\bm{J}$ & $\bm{\Delta}$ \\
\midrule
AI Scientist v1 & 51.8 [46.9, 56.6] & .48 [.37, .59] & .76 [.69, .83] \\
CoI-Agent & 3.8 [2.3, 6.1] & .93 [.85, .98] & .89 [.85, .93] \\
Si et al. & 96.0 [93.6, 97.5] & .04 [.02, .07] & .06 [.02, .10] \\
ARC & 0.0 [0.0, 1.0] & .22 [.21, .24] & .22 [.21, .24] \\
ResearchAgent & 98.5 [96.8, 99.3] & $-$.10 [$-$.14, $-$.05] & .21 [.15, .26] \\
ScholarEval & 0.0 [0.0, 1.0] & .00 & .00 \\
AI-Sci.-v2 & 39.2 [34.6, 44.1] & .56 [.35, .77] & .64 [.57, .72] \\
OpenNov.-style & 11.5 [8.7, 15.0] & .88 [.80, .96] & .90 [.85, .94] \\
\midrule
\textsc{ClaimNov} & 1.5 [0.7, 3.2] & .95 [.93, .98] & .92 [.88, .96] \\
\bottomrule
\end{tabular}}
\caption{95\% intervals for Table~\ref{tab:catchfa}. FA: Wilson. $J$, $\Delta$: cluster bootstrap by source and anchor.}
\label{tab:ci}
\end{table}

\begin{table}[!htbp]
\centering
\scriptsize
\setlength{\tabcolsep}{1.5pt}
\begin{tabular}{lccc}
\toprule
\rowcolor{gray!15}\textbf{Event} & \textbf{CoI-Agent} & \textbf{W/o profile} & \textbf{OpenNov.} \\
\midrule
Disguise caught & 10/11, $p{=}1.0$ & 5/11, $p{=}.21$ & 0/12, $p{<}.001$ \\
Src.\ out rejected & 10/23, $p{=}.035$ & 3/42, $p{<}10^{-9}$ & 8/28, $p{=}.001$ \\
Later-realized FA & 6/15, $p{=}.078$ & 4/45, $p{<}10^{-9}$ & 3/43, $p{<}10^{-9}$ \\
Claim set rejected & 0/12, $p{<}.001$ & 0/12, $p{<}.001$ & 0/8, $p{=}.008$ \\
Claim, $B$ on pass & 7/0, $p{=}.016$ & --- & --- \\
Hybrid passed & 20/5, $p{=}.004$ & 24/1, $p{<}10^{-5}$ & 15/7, $p{=}.13$ \\
Hybrid, $B$ on pass & 18/4, $p{=}.004$ & --- & --- \\
\bottomrule
\end{tabular}
\caption{Exact McNemar tests of \textsc{ClaimNov} against three checks, as discordant counts (\textsc{ClaimNov} only / other only).}
\label{tab:mcnemar}
\end{table}

\begin{table}[!htbp]
\centering
\scriptsize
\setlength{\tabcolsep}{3pt}
\begin{tabular}{>{\raggedright\arraybackslash}p{2.2cm}>{\raggedright\arraybackslash}p{5cm}}
\toprule
\rowcolor{gray!15}\textbf{Test} & \textbf{Result} \\
\midrule
Repeated verdict runs & 4 and 3 flips against the reported run, 1 between repeats; $J$ $+.941$, $+.941$, $+.938$; $\Delta$ $+.956$ in all three \\
Repeated CoI-Agent runs & 7 flips on each of 283 and 163 items \\
Paired $\Delta J$ (\textsc{ClaimNov} $-$ CoI-Agent) & $+.025$ $[-.05, +.12]$ on 474 items; $+.083$ $[-.04, +.25]$ on the 163 items with three CoI-Agent runs \\
\midrule
Clustered, vs CoI-Agent & $\Delta J$ $+.020$ $[-.04, +.11]$; source-out sign-flip permutation $p{=}.07$--$.09$ \\
Clustered, vs w/o profile & false-alarm rate $-.10$ $[-.18, -.03]$, $p{=}.002$; source-out $p{<}10^{-5}$ \\
Clustered, vs OpenNov.-style & false alarms and source-out $p{\le}.008$ \\
Anchors with false alarms & \textsc{ClaimNov} 3 (4 of 6 on one), CoI-Agent 2 (14 of 15 on one), w/o profile 11, OpenNov.-style 10 \\
\midrule
Endpoint & re-issuing the w/o-profile prompt on Bedrock raises its false alarms from 13 to 27 of 150 ideas, so its 11.8\% is a lower bound \\
\bottomrule
\end{tabular}
\caption{Run-to-run stability on 474 byte-identical items, and clustered tests.}
\label{tab:stability}
\end{table}

\begin{table}[!htbp]
\centering
\footnotesize
\setlength{\tabcolsep}{2.5pt}
\begin{tabular}{llcccc}
\toprule
\rowcolor{gray!15}\textbf{Set} & \textbf{Ranking} & \textbf{R@1} & \textbf{R@3} & \textbf{R@5} & \textbf{MRR} \\
\midrule
Claim & \textsc{ClaimNov} & .21 & .44 & .53 & .35 \\
(160) & dense & .37 & .59 & .72 & .52 \\
 & BM25 & .26 & .54 & .72 & .45 \\
 & OpenNov. & .24 & .26 & .26 & .25 \\
\midrule
Hybrid & \textsc{ClaimNov} & .25 & .39 & .42 & .32 \\
(100) & dense & .07 & .32 & .66 & .29 \\
 & BM25 & .74 & .92 & .96 & .84 \\
 & OpenNov. & .68 & .98 & .98 & .82 \\
\midrule
Disg. & \textsc{ClaimNov} & .89 & --- & .91 & .90 \\
(347) & dense & .87 & --- & .97 & .91 \\
 & BM25 & 1.00 & --- & 1.00 & 1.00 \\
\bottomrule
\end{tabular}
\caption{Rank of the relevant paper ($B$, or the source) in each ranking.}
\label{tab:ir}
\end{table}

\begin{table}[!htbp]
\centering
\footnotesize
\setlength{\tabcolsep}{4pt}
\begin{tabular}{llrrr}
\toprule
\rowcolor{gray!15}\textbf{Set} & \textbf{Level} & \textbf{Papers} & \textbf{Relevant} & \textbf{Prec.} \\
\midrule
Claim (160) & $\ge$3 & 291 & 54 & .19 \\
Claim (160) & $\ge$4 & 14 & 2 & .14 \\
Hybrid (100) & $\ge$4 & 15 & 13 & .87 \\
Disg.\ (347) & $\ge$4 & 326 & 303 & .93 \\
\bottomrule
\end{tabular}
\caption{Profile papers at the top coverage levels, and how many are the relevant paper. Level 4 or above fires for 9 of 400 later-realized ideas.}
\label{tab:levels}
\end{table}

\FloatBarrier
\section{Robustness}
\label{app:robust}

In this section, we test some assumptions to better understand if our main results depend on our specific labels, judge model, pool sizes, or the particular data we collected.

Because our later-realized labels come from one Opus labeler, we recompute false alarms under three alternative labelers in Table~\ref{tab:alt-labels}, specifically (1) a labeler from a different model family (gpt-5), (2) the ideas both Opus and gpt-5 agree on, and (3) the full set without the $n=101$ ideas a whole-idea Opus labeling pass considered previously done. For all of these labeling paradigms, \textsc{ClaimNov}'s false-alarm rate stays between 1.3\% and 1.5\%.

We check whether the claim-level novelty profile we provide for our final verdict is only helpful to our particular judge (Sonnet~4.6) by rerunning this stage with two non-Anthropic judges, specifically gpt-oss-120b (117B parameters) and Llama~4 Maverick (17B active, 400B total) in Table~\ref{tab:nonclaude}. These are weaker models, so catch does drop, but we still observe as with Sonnet that providing the profile cuts false alarms and source-out rejections relative to the base judge reading nearest abstracts.

We check in Table~\ref{tab:baselines} whether baselines can match the judged checks. These are thresholds on lexical or embedding similarity, a judge that sees only the single nearest paper, and the extracted score-based checks at their best threshold. The baselines are not competitive with \textsc{ClaimNov}'s $J$.

Because our disguise pools hold only 20 papers, in Table~\ref{tab:largepools} we artificially increase the pool size to $\approx 200$ to investigate if our catch can survive more realistic pool sizes. While catch does fall on the hardest rung, source-out rejections stay near zero.

We also perform a handful of smaller checks that further support these findings (Table~\ref{tab:robust-misc}). Asked with no pool, Sonnet~4.6 names $B$ first for at most 1 of the 160 claim-disputed proposals, so recall of pre-$T$ papers does not drive that set. Appendix~\ref{app:audit} reports a human audit of 60 generated ideas.

\begin{table}[!htbp]
\centering
\footnotesize
\setlength{\tabcolsep}{3pt}
\begin{tabular}{lrr}
\toprule
\rowcolor{gray!15}\textbf{Check} & \textbf{\textsc{ClaimNov}} & \textbf{CoI} \\
\midrule
\multicolumn{3}{l}{\emph{All 400 disguises}} \\
\quad caught & 386 & 388 \\
\quad src.\ out rejected & 18 & 31 \\
\multicolumn{3}{l}{\emph{Pilot items excluded}} \\
\quad catch (340) & 96.5\% & --- \\
\quad FA (282) & 0.7\% & 5.3\% \\
\quad accepted rejected & 2/51 & --- \\
\multicolumn{3}{l}{\emph{Other}} \\
\quad reworded question, FA & 6, 8 & --- \\
\quad anchor in pool, FA & 7 & 3 \\
\quad all 3{,}000 ideas, rejected & 2.9\% & --- \\
\quad hybrids, base passes, kept & 91/99 & 74/98 \\
\bottomrule
\end{tabular}
\caption{Further robustness checks. Pilot items excluded: items used to choose the verdict question, judge tier and thresholds. CoI: CoI-Agent.}
\label{tab:robust-misc}
\end{table}

\paragraph{Reproducibility.} We release all code and prompts at \url{https://github.com/emmatliu/claimnov}, including the scripts that build each dataset and run each check, together with the per-item records behind our tables. We release the generated ideas themselves. For source papers, pool papers and ICLR reviews, we release identifiers and titles rather than paper or review text, and the released code re-mines the claim disputes from public OpenReview reviews. We do not release the disguised or hybrid proposals because they restate published papers, but we do release code to rebuild them.

\begin{table}[!htbp]
\centering
\footnotesize
\setlength{\tabcolsep}{3pt}
\begin{tabular}{lrrrr}
\toprule
\rowcolor{gray!15}\textbf{Check} & \textbf{All} & \textbf{gpt-5} & \textbf{Agree} & \textbf{$-$101} \\
\rowcolor{gray!15} & \textbf{(400)} & \textbf{(305)} & \textbf{(292)} & \textbf{(299)} \\
\midrule
AI Scientist v1 & 51.8 & 42.6 & --- & 38.8 \\
CoI-Agent & 3.8 & 0.0 & 0.0 & 1.3 \\
Si et al. & 96.0 & 95.1 & --- & 94.6 \\
ResearchAgent & 98.5 & 98.7 & --- & --- \\
AI-Sci.-v2 & 39.2 & 29.5 & --- & --- \\
OpenNov.-style & 11.5 & 10.5 & --- & 11.7 \\
W/o profile & 11.8 & 7.2 & --- & 7.4 \\
\midrule
\textsc{ClaimNov} & 1.5 & 1.3 & 1.4 & 1.3 \\
\bottomrule
\end{tabular}
\caption{False alarms (\%) on later-realized ideas under alternative labels. gpt-5: ideas a gpt-5 labeler confirms as novel. Agree: Opus and gpt-5 agree. $-$101: excluding 101 ideas a whole-idea pass called already done.}
\label{tab:alt-labels}
\end{table}

\begin{table}[!htbp]
\centering
\footnotesize
\setlength{\tabcolsep}{3pt}
\begin{tabular}{lrrrr}
\toprule
 & \multicolumn{2}{c}{\textbf{gpt-oss-120b}} & \multicolumn{2}{c}{\textbf{Llama~4 Maverick}} \\
\cmidrule(lr){2-3}\cmidrule(lr){4-5}
\rowcolor{gray!15}\textbf{Measure} & \textbf{Profile} & \textbf{Abstr.} & \textbf{Profile} & \textbf{Abstr.} \\
\midrule
Subset FA (341) & 0 & 15 & 25 & 91 \\
Subset src.\ out (65) & 0 & 7 & 6 & 38 \\
Subset catch (68) & 56 & 60 & --- & --- \\
\midrule
Full FA (400) & 1 & 25 & 26 & 104 \\
Full src.\ out (347) & 7 & 31 & 30 & 175 \\
Full catch (347) & 287 & 319 & 304 & 343 \\
Full $J$ & $+.83$ & $+.86$ & $+.81$ & $+.73$ \\
\bottomrule
\end{tabular}
\caption{Non-Claude verdict judges. Profile: the judge reads the claim profile. Abstr.: it reads the top-8 abstracts. Subset: prompts byte-identical to the Sonnet~4.6 runs.}
\label{tab:nonclaude}
\end{table}

\begin{table}[!htbp]
\centering
\scriptsize
\setlength{\tabcolsep}{2.5pt}
\begin{tabular}{lrrrrr}
\toprule
\rowcolor{gray!15}\textbf{Baseline} & \textbf{Catch} & \textbf{Src.\ out} & \textbf{FA} & $\bm{J}$ & \textbf{Hyb.} \\
\midrule
BM25, max score & --- & --- & --- & $+.57$ & --- \\
Embedding, max cosine & --- & --- & --- & $+.21$ & --- \\
SciNCL, max cosine & --- & --- & --- & $+.45$ & --- \\
SPECTER2, max cosine & --- & --- & --- & $+.33$ & --- \\
Retrieve-and-judge & 378 & 65 & 47 & $+.88$ & 39 \\
ScholarEval, best thr. & 359 & 134 & 29.5\% & $+.65$ & --- \\
\midrule
\textsc{ClaimNov} & 367 & 18 & 6 & $+.95$ & 91 \\
\bottomrule
\end{tabular}
\caption{Simpler baselines on the full sets. Similarity thresholds are cross-validated, and ScholarEval's is the best threshold on the evaluation data.}
\label{tab:baselines}
\end{table}

\begin{table}[!htbp]
\centering
\footnotesize
\setlength{\tabcolsep}{4pt}
\begin{tabular}{lrrrr}
\toprule
 & \multicolumn{2}{c}{\textbf{\textsc{ClaimNov}}} & \multicolumn{2}{c}{\textbf{CoI-Agent}} \\
\cmidrule(lr){2-3}\cmidrule(lr){4-5}
\rowcolor{gray!15}\textbf{Rung} & \textbf{Catch} & \textbf{Src.\ out} & \textbf{Catch} & \textbf{Src.\ out} \\
\midrule
r1 & 40 $\to$ 40 & 0 & --- & --- \\
r2 & 40 $\to$ 38 & 0 & --- & --- \\
r3 & 39 $\to$ 35 & 1 & 35 & 5 \\
r4 & 35 $\to$ 28 & 1 & 35 & 8 \\
\bottomrule
\end{tabular}
\caption{Catch with pools enlarged from 20 to about 200 papers (40 sources). CoI-Agent is run on rungs 3 and 4 only.}
\label{tab:largepools}
\end{table}
\FloatBarrier
\section{Natural Already-Done Ideas}
\label{app:natural}
A whole-idea Opus sorter, reading the eight nearest pool papers, matched 837 of the 3{,}000 generated ideas to a paper that already does the idea. The claim-level labeler overruled it on 101 (these are in the later-realized set), and 407 of the remaining 736 have the shipped profile (Table~\ref{tab:natural}). The set is noisier than the disguises because the sorter read exactly the abstracts the w/o-profile ablation reads. Appendix~\ref{app:audit} reports a human audit of 15 of these ideas.

\begin{table}[!htbp]
\centering
\footnotesize
\setlength{\tabcolsep}{3pt}
\begin{tabular}{lrrr}
\toprule
\rowcolor{gray!15}\textbf{Measure} & \textbf{W/o profile} & \textbf{+ abstr.} & \textbf{\textsc{ClaimNov}} \\
\midrule
Rejects (407) & 224 (55.0\%) & 145 (35.6\%) & 72 (17.7\%) \\
Catch (347) & --- & 345 & 334 \\
FA (400) & --- & 25 & 6 \\
Hybrids kept & --- & 77 & 91 \\
Claim rej.\ (158) & --- & 12 & 5 \\
\bottomrule
\end{tabular}
\caption{Sorter-matched ideas and the main sets. + abstr.: the verdict reads the profile and the eight nearest abstracts.}
\label{tab:natural}
\end{table}

\FloatBarrier
\section{In-the-Loop Pilot}
\label{app:loop}
We plug each check into a generate--check--regenerate loop on two anchors (In-Context Edit and MoBA, 178--185-paper pools, 20 ideas each). Every arm starts from the same first idea, and a rejection returns its reason and cited paper to the generator, with up to three regenerations (Table~\ref{tab:loop}). A pairwise review-score proxy \citep{si2025llms} separates committed from unchecked ideas on only 3 seeds for \textsc{ClaimNov} and 1 for CoI-Agent, too few to compare the arms, so we leave in-the-loop evaluation to future work.

\begin{table}[!htbp]
\centering
\footnotesize
\setlength{\tabcolsep}{3pt}
\begin{tabular}{lcccc}
\toprule
\rowcolor{gray!15}\textbf{Arm} & \textbf{1st rej.} & \textbf{Checks} & \textbf{\$} & \textbf{Seconds} \\
\midrule
\textsc{ClaimNov} & 3/40 & 1.08 & 0.087 & 13.8 \\
CoI-Agent & 1/40 & 1.03 & 0.047 & 22.2 \\
No check & --- & 0 & 0 & 0 \\
\bottomrule
\end{tabular}
\caption{In-the-loop pilot on 40 ideas, per committed idea.}
\label{tab:loop}
\end{table}

\FloatBarrier
\section{Case Study}
\label{app:case}
In Table~\ref{fig:case-trace}, we trace two proposals through \textsc{ClaimNov}. The first rewrites the ICLR 2022 Spotlight \emph{Amortized Tree Generation for Bottom-up Synthesis Planning and Synthesizable Molecular Design}, whose reviewer attributed its MDP formulation to \emph{Barking up the right tree}, with that paper in the pool. The second is a rung-4 disguise of \emph{Ratio-Variance Regularized Policy Optimization} with the source in the pool.

\begin{table*}[!htbp]
\footnotesize
\setlength{\tabcolsep}{3pt}
\renewcommand{\arraystretch}{1.1}
\begin{minipage}[t]{0.485\textwidth}
\textbf{Claim-level: ICLR Spotlight, one disputed claim.}
$N(P)=0.71$ $\Rightarrow$ \verdictbox{okgreenbg}{okgreen}{pass}\\[2pt]
\begin{tabular}{l>{\raggedright\arraybackslash}p{2.45cm}>{\raggedright\arraybackslash}p{2.75cm}r}
\toprule
\rowcolor{gray!15}\textbf{Dim.} & \textbf{Claim} & \textbf{Top matches (level)} & $\bm{\nu_c}$ \\
\midrule
\dimtag{problem} & Molecular design and synthesis planning as one task of conditional pathway generation & 2 Synthesizability of generated molecules & 0.75 \\
\dimtag{method} & Amortized generation of synthetic pathways as an MDP conditioned on a target embedding & \src{3 Barking up the right tree (RetroDog)}; 2 A model to search for synthesizable molecules; 2 DOGS & 0.62 \\
\bottomrule
\end{tabular}\\[3pt]
\textbf{Judge:} \emph{``No single paper in the profile proposes the same
unified framework \ldots\ The closest works (Barking up the right tree,
A Model to Search for Synthesizable Molecules) share partial
methodological overlap but do not combine the full set of
contributions.''}
\end{minipage}\hfill
\begin{minipage}[t]{0.485\textwidth}
\textbf{Whole-proposal: rung-4 disguise, source planted.}
$N(P)=0.52$ $\Rightarrow$ \verdictbox{badredbg}{badred}{reject}\\[2pt]
\begin{tabular}{l>{\raggedright\arraybackslash}p{2.45cm}>{\raggedright\arraybackslash}p{2.75cm}r}
\toprule
\rowcolor{gray!15}\textbf{Dim.} & \textbf{Claim} & \textbf{Top matches (level)} & $\bm{\nu_c}$ \\
\midrule
\dimtag{method} & Distributional-control approach to stability and efficiency in policy gradients & \src{4 Ratio-Variance Reg.\ PO}; 3 Differentiable trust-region layers; 3 Off-policy TRPO & 0.39 \\
\dimtag{method} & Soft constraint on the second moment of the policy ratio as a trust-region proxy & \src{4 Ratio-Variance Reg.\ PO}; 3 Stochastic trust-region framework; 3 Two-phase KL penalization & 0.39 \\
\dimtag{result} & Substantial gains in sample efficiency and final performance over PPO & \src{4 Ratio-Variance Reg.\ PO}; 3 Quasi-Newton TRPO; 3 Tapered off-policy REINFORCE & 0.38 \\
\dimtag{problem} & Evaluate across LM reasoning and robotic control & 2 Metric-aware trust regions (1 retrieved) & 0.75 \\
\bottomrule
\end{tabular}\\[3pt]
\textbf{Judge:} not novel; establishing paper \src{Ratio-Variance Reg.\ PO};
established claims 1--3. \emph{``Already proposes the same core idea:
constraining policy ratio variance as a principled soft alternative to
hard clipping, implemented via a primal-dual optimization framework
\ldots\ at 4/5 coverage across all core claims.''} Source removed:
\verdictbox{okgreenbg}{okgreen}{pass}.
\end{minipage}
\caption{\textsc{ClaimNov} on a claim-level dispute (left), which it passes, and a rung-4 disguise (right), which it rejects. The reviewer's paper and the source are in \src{blue}.}
\label{fig:case-trace}
\end{table*}

\section{Component Validations}
\label{app:claim-validation}
We explain the validations we performed on each component in \S\ref{sec:claim-module}.

\subsection{Claim extraction}
We extract typed claims from the abstract with \judgemodel{} (temperature 0, JSON mode) using a GEPA-tuned prompt, scored against a 20-paper, three-annotator ground truth (Table~\ref{tab:val-extraction}). When we extract from introductions rather than abstracts, F1 drops from 0.577 to 0.490, so in \textsc{ClaimNov} we use abstracts. gpt-5-mini has equivalent performance on extraction but performs poorly as the entailment judge (Table~\ref{tab:tiersweep}).

\begin{table}[!htbp]
\centering
\footnotesize
\setlength{\tabcolsep}{2pt}
\begin{tabular}{llccc}
\toprule
\rowcolor{gray!15}\textbf{Prompt} & \textbf{Model} & \textbf{F1} & \textbf{dim.\ $\bm{\kappa}$} & \textbf{cov.} \\
\midrule
Human pair & --- & .543 & .515$^\dagger$ & --- \\
\midrule
pre-GEPA & \judgemodel{} & .646 / .667 & .596 / .662 & 41--42 \\
\textbf{GEPA} & \textbf{\judgemodel{}} & \textbf{.674 / .695} & \textbf{.793 / .744} & 38--39 \\
GEPA & gpt-5-nano & .642 & .696 & 41 \\
GEPA & gpt-5-mini & .673 & .742 & 43 \\
\bottomrule
\end{tabular}
\caption{Extraction against a 20-paper human ground truth (two runs per configuration). Bold: shipped. $\dagger$ Fleiss' $\kappa$. cov.: reviewer-named contributions covered, of 75.}
\label{tab:val-extraction}
\end{table}

\subsection{Same-dimension retrieval}
To validate same-dimension retrieval, we embed each claim's title plus its attribute--value text with \embmodel{}, then rank its cosine similarity with same-dimension pool claims. In Table~\ref{tab:retrieval} we score recall of the reviewer-named work on 150 ICLR 2022--23 reviewer-tell pairs with $A$'s own references as distractors, and in Table~\ref{tab:val-recall} we look at the value of staying within the same dimension. Because we find that only 3 of the 159 disputes are not in same dimension, we keep same dimension retrieval rather than retrieving over all types of claims.

\begin{table}[!htbp]
\centering
\small
\begin{tabular}{lrccc}
\toprule
\rowcolor{gray!15}\textbf{Bucket} & $\bm{n}$ & \textbf{@5} & \textbf{@10} & \textbf{@13} \\
\midrule
Overall & 150 & 0.55 & \textbf{0.79} & \textbf{0.90} \\
\midrule
Problem & 39 & 0.62 & 0.87 & 0.95 \\
Methodology & 40 & 0.62 & 0.80 & 0.93 \\
Resource & 28 & 0.57 & 0.82 & 0.93 \\
Result & 43 & 0.42 & 0.67 & 0.81 \\
\midrule
$B$ cited by $A$ & 86 & 0.50 & 0.79 & 0.90 \\
$B$ uncited & 47 & \textbf{0.66} & 0.77 & 0.87 \\
\bottomrule
\end{tabular}
\caption{Recall of reviewer-named prior work, with the target paper's references as distractors.}
\label{tab:retrieval}
\end{table}

\begin{table}[!htbp]
\centering
\small
\setlength{\tabcolsep}{3pt}
\begin{tabular}{lrcccc}
\toprule
\rowcolor{gray!15}\textbf{Configuration} & \textbf{pool} & \textbf{@1} & \textbf{@5} & \textbf{@10} & \textbf{@13} \\
\midrule
D1 claim only, same-dim & 19.7 & .21 & .56 & .75 & .87 \\
D2 D1 + title on query & 19.7 & .21 & .59 & .82 & .89 \\
D3 D1 + title on candidates & 19.7 & .21 & .54 & .78 & .85 \\
\textbf{D4 title on both (shipped)} & 19.7 & \textbf{.26} & \textbf{.64} & \textbf{.84} & \textbf{.90} \\
D5 D3, any-dimension pool & 75.4 & .21 & .41 & .52 & .60 \\
\midrule
Oracle: reviewer sentence & 19.7 & .28 & .57 & .80 & .89 \\
Oracle + facet, titled cands. & 19.7 & .32 & .69 & .86 & .89 \\
\bottomrule
\end{tabular}
\caption{Recall of reviewer-named prior work by query and candidate representation.}
\label{tab:val-recall}
\end{table}

\paragraph{Retrieval depth.}
\label{app:kdepth}
From above validation we set depth as $k = \max(10, 0.2\, n_d)$ capped at 40. Table~\ref{tab:kdepth} checks this heuristic on the $n=837$ sorter-matched ideas (pools of up to 400 papers, median 174). In this set, using this depth rule means that only 5 of the 837 matched papers are never retrieved.

\begin{table}[!htbp]
\centering
\small
\setlength{\tabcolsep}{3pt}
\begin{tabularx}{\linewidth}{lRRRRRR}
\toprule
\rowcolor{gray!15}\textbf{Fixed} $\bm{k}$ & 5 & 10 & 20 & 30 & 40 & 80 \\
\midrule
In profile & 44\% & 57\% & 69\% & 77\% & 82\% & 91\% \\
\midrule
\rowcolor{gray!15}\textbf{Share of cands.} & 5\% & 10\% & 15\% & 20\% & 30\% & \\
\midrule
In profile & 46\% & 60\% & 69\% & 75\% & 83\% & \\
\bottomrule
\end{tabularx}
\caption{Share of sorter-matched ideas whose matched paper enters the profile, by retrieval depth.}
\label{tab:kdepth}
\end{table}

\subsection{Graded entailment judge}
\label{app:tiers}
In Tables~\ref{tab:val-context}--\ref{tab:val-p2}, we evaluate and calibrate the graded entailment stage of \textsc{ClaimNov}. Table~\ref{tab:val-context} compares the value of added context (specifically, titles and abstracts) to the value of higher model tiers; we find that context is comparatively more valuable. Table~\ref{tab:tiersweep} investigates if the graded score alone can separate reviewer-attested novel claims ($n=66$) from reviewer-attested established claims ($n=159$) across six judge tiers and three retrieval settings, but we find that embedding-only similarity is a strong baseline here. Table~\ref{tab:val-human} compares the judge's use of the 1--5 entailment scale with that of three human annotators on $n=50$ pairs; we find every tier agrees with the human majority more than the annotators agree with each other. Table~\ref{tab:val-p2} tests whether our graded score can successfully rank a claim a reviewer specifically said was novel as higher than the sibling claims from the same paper ($n=86$ papers and $n=233$ claims); we find that it does so above chance, 59\% of the time. All in all, we observe that the graded score can order claims well, but that there isn't a good way to set a pass/fail novelty threshold based off the graded score.

\begin{table}[!htbp]
\centering
\small
\setlength{\tabcolsep}{4pt}
\begin{tabular}{llcc}
\toprule
\rowcolor{gray!15}\textbf{Prompt} & \textbf{Judge} & \textbf{gold mean} & \textbf{separation} \\
\midrule
facets only & \judgemodel{} & 2.43 & 0.57 \\
+ title, abstract & \judgemodel{} & 2.70 & 0.93 \\
+ title, abstract & Haiku 4.5 & 2.70 & 0.97 \\
+ title, abstract & Sonnet 4.6 & 2.87 & 1.03 \\
+ title, abstract & Opus 4.6 & 3.07 & 1.07 \\
\midrule
+ title, abstract & gpt-5-nano & 2.97 & 0.87 \\
+ title, abstract & gpt-5-mini & 2.80 & 0.67 \\
\bottomrule
\end{tabular}
\caption{Prompt context versus judge tier (30 gold and 30 distractor pairs).}
\label{tab:val-context}
\end{table}

\begin{table}[!htbp]
\centering
\small
\begin{tabular}{lrrr}
\toprule
\rowcolor{gray!15}\textbf{Judge} & \textbf{base} & \textbf{titled} & \textbf{titled} \\
\rowcolor{gray!15} & $k{=}5$ & $k{=}5$ & $k{=}10$ \\
\midrule
gpt-5-nano & .460 & .451 & .440 \\
gpt-5-mini & .438 & .503 & .443 \\
gpt-4o-mini & .516 & .551 & .546 \\
Haiku 4.5 & .539 & .564 & .527 \\
Sonnet 4.6 & .589 & .606 & .611 \\
Opus 4.6 & .576 & .601 & .594 \\
\midrule
embedding-only & .607 & --- & --- \\
\bottomrule
\end{tabular}
\caption{AUC of the graded score separating novel from established claims. Titled: the shipped retrieval representation.}
\label{tab:tiersweep}
\end{table}

\begin{table}[!htbp]
\centering
\footnotesize
\setlength{\tabcolsep}{3pt}
\begin{tabular}{llcccc}
\toprule
\rowcolor{gray!15}\textbf{Judge} & \textbf{prompt} & $\bm{\alpha}$ & \textbf{MAD} & \textbf{exact} & \textbf{$\pm$1} \\
\midrule
Human pair & --- & .42 & .95 & .41 & .74 \\
\midrule
\judgemodel{} & claim & .73 & .62 & .43 & .95 \\
\judgemodel{} & + titles & .73 & .57 & .49 & .95 \\
Haiku 4.5 & claim & \textbf{.86} & \textbf{.35} & \textbf{.68} & \textbf{.97} \\
Haiku 4.5 & + titles & .84 & .38 & .65 & .97 \\
Sonnet 4.6 & claim & .78 & .49 & .57 & .95 \\
Sonnet 4.6 & + titles & .75 & .54 & .51 & .95 \\
Opus 4.6 & claim & .74 & .54 & .51 & .95 \\
Opus 4.6 & + titles & .57 & .73 & .35 & .92 \\
\bottomrule
\end{tabular}
\caption{Graded entailment against the human majority label (37 pairs). The human row is agreement between annotators.}
\label{tab:val-human}
\end{table}

\begin{table}[!htbp]
\centering
\footnotesize
\setlength{\tabcolsep}{2pt}
\begin{tabular}{lccc}
\toprule
\rowcolor{gray!15}\textbf{Scorer} & \textbf{win rate} & \textbf{95\% CI} & \textbf{claim FA} \\
\midrule
Graded, \judgemodel{} & 0.591 & [0.525, 0.655] & --- \\
Binary veto, Sonnet 4.6 & 0.500 & floor & 0 / 233 \\
\bottomrule
\end{tabular}
\caption{Win rate of a reviewer-praised claim over its sibling claims.}
\label{tab:val-p2}
\end{table}

\subsection{Verdict}
\paragraph{Calibration.}
We use $n=20$ disputed pairs, $n=32$ later-realized decoys, and $n=106$ already-done ideas to try and calibrate a particular novel or non-novel threshold for $N(P)$, as we show in Table~\ref{tab:val-calibration}. We consider a knockout entailment level, relation labels, and a per-candidate question; the verdict stage we ultimately adopted considers novelty once over the whole profile (\S\ref{sec:verdict}), while the per-candidate question is the per-claim ablation of Table~\ref{tab:catchfa}.

\begin{table}[!htbp]
\centering
\footnotesize
\setlength{\tabcolsep}{2pt}
\begin{tabular}{llccc}
\toprule
\rowcolor{gray!15}\textbf{Rule} & \textbf{Judge} & \textbf{N2} & \textbf{(a)} & \textbf{(b) FA} \\
\midrule
$t{=}.5$, ko$=$4$^\ast$ & 4o-mini & 0\% & 0\% & 1\% \\
$t{=}.6$, no ko & 4o-mini & 15\% & 46\% & 30\% \\
ko$=$3, any $t$ & 4o-mini & 75\% & 82\% & 78\% \\
$t{=}.5$, ko$=$4 & Sonnet 4.6 & 10\% & --- & 3\% \\
ko$=$3, any $t$ & Sonnet 4.6 & 90\% & --- & 72\% \\
\midrule
rel.\ same/subsumes & 4o-mini & 5\% & --- & --- \\
rel.\ $+$ partial & 4o-mini & --- & --- & 100\% \\
rel.\ same/subsumes & Sonnet 4.6 & 0\% & --- & --- \\
\midrule
veto & 4o-mini & 45\% & 60\% & 56\% \\
veto & Haiku 4.5 & 20\% & --- & 0\% \\
veto & Sonnet 4.6 & 35\% & --- & 0\% \\
veto & Opus 4.6 & 15\% & --- & 0\% \\
\midrule
veto, outcome-reworded & Sonnet 4.6 & 10\% & --- & 0\% \\
\bottomrule
\end{tabular}
\caption{Pilot calibration of decision rules ($t$: threshold on $N(P)$, ko: knockout level, $\ast$: shipped). N2: catch on 20 disputed pairs. (a): catch on 106 already-done ideas. (b) FA: false alarms on 32 decoys.}
\label{tab:val-calibration}
\end{table}

\section{AI Disclosure Statement}
LLM agents (e.g., Claude Code) were used to assist with experiment and analysis code, to advise on paper structure and clarity, to generate results tables and draft appendix text, and to prepare released data and code files. All scientific contributions, including the \textsc{ClaimNov} methodology, experimental design, and data collection approaches, were devised by the authors.

\end{document}